\documentclass{article}

\PassOptionsToPackage{numbers, compress}{natbib}

\usepackage[preprint]{neurips_2024}

\usepackage[utf8]{inputenc} 
\usepackage[T1]{fontenc}    
\usepackage{hyperref}       
\usepackage{algorithm}
\usepackage{algorithmicx}
\usepackage{algpseudocode}
\usepackage{url}            
\usepackage{booktabs}       
\usepackage{amsfonts}       
\usepackage{nicefrac}       
\usepackage{microtype}      
\usepackage{xcolor}         

\usepackage{amsmath}
\usepackage{amssymb}
\usepackage{mathtools}
\usepackage{amsthm}
\usepackage{graphicx} 
\usepackage{subcaption}
\usepackage{multirow}
\usepackage{afterpage}

\title{Rate-Adaptive Quantization: \\ A Multi-Rate Codebook Adaptation for \\ Vector Quantization-based Generative Models}

\author{%
  Jiwan~Seo,  \ \ \ Joonhyuk~Kang\thanks{Corresponding author}\\
  Department of Electrical Engineering \\
  Korea Advanced Institute of Science and Technology (KAIST)\\
  \texttt{jeewan0516@kaist.ac.kr}, \ \ \ \texttt{jhkang@ee.kaist.ac.kr} \\
}

\begin{document}

\maketitle

\begin{abstract}
    Learning discrete representations with vector quantization (VQ) has emerged as a powerful approach in various generative models. However, most VQ-based models rely on a single, fixed-rate codebook, requiring extensive retraining for new bitrates or efficiency requirements. We introduce \emph{Rate-Adaptive Quantization (RAQ)}, a multi-rate codebook adaptation framework for VQ-based generative models. RAQ applies a data-driven approach to generate variable-rate codebooks from a single baseline VQ model, enabling flexible tradeoffs between compression and reconstruction fidelity. Additionally, we provide a simple clustering-based procedure for pre-trained VQ models, offering an alternative when retraining is infeasible. Our experiments show that RAQ performs effectively across multiple rates, often outperforming conventional fixed-rate VQ baselines. By enabling a single system to seamlessly handle diverse bitrate requirements, RAQ extends the adaptability of VQ-based generative models and broadens their applicability to data compression, reconstruction, and generation tasks.
\end{abstract}

\section{Introduction}
Vector quantization (VQ) \citep{gray1984vector} is a fundamental technique for learning discrete representations for various tasks \citep{krishnamurthy1990neural, gong2014compressing, van2020vector} in the field of machine learning. The Vector Quantized Variational AutoEncoder (VQ-VAE) \citep{van2017neural, razavi2019generating}, which extends the encoder-decoder structure of the Variational Autoencoder (VAE) \citep{kingma2013auto, rezende2018taming}, introduces discrete latent representations that have proven effective across vision \citep{razavi2019generating, esser2021taming}, audio \citep{dhariwal2020jukebox, yang2023diffsound}, and speech tasks \citep{kumar2019melgan, xing2023codetalker}. The inherently discrete nature of these modalities makes VQ particularly well-suited for complex inference and generation.

Recent developments have further enhanced VQ-based discrete representation learning by integrating it with deep generative models, such as Generative Adversarial Networks (GANs) \citep{esser2021taming} and Denoising Diffusion Probabilistic Models (DDPMs) \citep{cohen2022diffusion, gu2022vector, yang2023diffsound}. As VQ-based generative models are integrated into these diverse generative frameworks, their utility and applicability in various tasks are becoming increasingly evident. However, even with these advancements, \emph{scalability} remains a bottleneck. In practical settings such as live streaming, telepresence, and on-device applications, the available bandwidth and compute resources can fluctuate dramatically. A single fixed-rate VQ model either wastes bits when higher quality is possible or severely degrades fidelity under tight constraints. Maintaining separate VQ models for each bitrate is infeasible and incurs significant overhead. Hence, a robust framework that can seamlessly adapt its compression rate is crucial for real-world deployments.

Several works have explored enhancing the flexibility of codebooks. \citet{li2023resizing} introduced a codebook-resizing technique for publicly available VQ models by applying hyperbolic embeddings, \citet{malka2023learning} propose a nested codebook to support multiple quantization levels, and \citet{guo2023predicting} explore multi-codebook vector quantization in a knowledge distillation setting. Recently, \citet{huijben2024residual} focused on unsupervised codebook generation based on residual quantization by studying the vector quantizer itself. While these methods enrich representation capacity, they typically require extensive changes to previous VQ architectures or reduce the resolution of quantized feature maps \citep{guo2022msmc}, making them less practical for real-world applications.

In this paper, we present \emph{Rate-Adaptive Quantization (RAQ)}, a framework designed to flexibly modulate the effective codebook size of a single VQ-based model without retraining. By incorporating a Sequence-to-Sequence (Seq2Seq) \citep{sutskever2014sequence} into the VQ-based generative model, RAQ enables one system to cover multiple compression levels, reducing the need for separate models dedicated to each rate. This adaptability not only minimizes storage and maintenance costs but also provides a smoother user experience in real-time communications or streaming environments, where bandwidth availability can vary from moment to moment. While our main focus is on the Seq2Seq-based RAQ, we additionally propose a model-based alternative approach that utilizes differential $k$-means (DKM) \citep{cho2021dkm} clustering on a pre-trained VQ model, allowing codebook adaptation without the need for new parameters or retraining. This simple approach provides a practical fallback in scenarios where retraining or model modification is not feasible.

Our contributions are summarized as follows:
\begin{itemize}
    \item We propose the \emph{Rate-Adaptive Quantization (RAQ)} framework for flexible multi-rate codebook adaptation using a Sequence-to-Sequence (Seq2Seq) model. This method can be integrated into existing VQ-based generative models with minimal modifications.
    \item To mitigate distribution mismatch in autoregressive codebook generation, we introduce a \emph{cross-forcing} training procedure. This approach ensures stable codebook generation across diverse rates and enhances reconstruction fidelity.
    \item We evaluate RAQ on several VQ-based benchmarks and show that a \textit{single} RAQ-enabled model consistently meets or exceeds the performance of multiple fixed-rate VQ baselines while using the same encoder-decoder architecture.
\end{itemize}

\section{Background}

\subsection{Vector-Quantized Variational AutoEncoder}

VQ-VAEs \citep{van2017neural} can successfully represent meaningful features that span multiple dimensions of data space by discretizing continuous latent variables to the nearest codebook vector in the codebook. In VQ-VAE, learning of discrete representations is achieved by quantizing the encoded latent variables to their nearest neighbors in a trainable codebook and decoding the input data from the discrete latent variables. To represent the data $\mathbf{x}$ from dataset $\mathcal{D}$ discretely, a codebook $\mathbf{e}$ consisting of $K$ learnable codebook vectors $\{e_i\}_{i=1}^K \subset \mathbb{R}^d$ is employed (where $d$ denotes the dimensionality of each codebook vector $e_i$). The quantized discrete latent variable $\mathbf{z}_q(\mathbf{x}|\mathbf{e})$ is decoded to reconstruct the data $\mathbf{x}$. The vector quantizer $Q$ modeled as deterministic categorical posterior maps each spatial position $[m,n]$ of the continuous latent representation $\mathbf{z}_{e}(\mathbf{x})[m, n]$ of the data $\mathbf{x}$ by a deterministic encoder $f_{\phi}$ to $\mathbf{z}_q(\mathbf{x}|\mathbf{e}) [m, n]$ by finding the nearest neighbor from the codebook $\mathbf{e}=\{e_i\}_{i=1}^K$ as
\begin{equation}
 \mathbf{z}_q(\mathbf{x}|\mathbf{e})[m, n] = Q\Big(\mathbf{z}_{e}(\mathbf{x})[m, n]\big|\mathbf{e}\Big) = {\arg \min }_{i}\left\| \mathbf{z}_{e}(\mathbf{x})[m, n]-e_i \right\|,
\label{eq_quantization}   
\end{equation}
The quantized representation uses $\log_2{K}$ bits to index one of the $K$ selected codebook vectors $\{e_i\}_{i=1}^K$. The deterministic decoder $f_\theta$ reconstructs the data $\mathbf{x}$ from the quantized discrete latent variable $\mathbf{z}_q(\mathbf{x}|\mathbf{e})$ as $\hat{\mathbf{x}} = f_{\theta}\big(\mathbf{z}_q(\mathbf{x}|\mathbf{e})|\mathbf{e})\big)$. During the training process, the encoder $f_{\phi}$, decoder $f_{\theta}$, and codebook $\mathbf{e}$ are jointly optimized to minimize the loss $\mathcal{L}_{\text{VQ}}\big(\phi,\theta, \mathbf{e};\mathbf{x}\big)=$
\begin{equation}
\underbrace{\log p_{\theta}(\mathbf{x}|\mathbf{z}_q(\mathbf{x}|\mathbf{e}))}_{\mathcal{L}_{\mathrm{recon}}} + \underbrace{\big|\big|\text{sg}\big[f_{\phi}(\mathbf{x})\big]-\mathbf{z}_q(\mathbf{x}|\mathbf{e})\big|\big|_2^2}_{\mathcal{L}_{\mathrm{embed}}} + \underbrace{\beta\big|\big|\text{sg}\big[\mathbf{z}_q(\mathbf{x}|\mathbf{e})\big] - f_{\phi}(\mathbf{x})\big|\big|_2^2}_{\mathcal{L}_{\mathrm{commit}}}  
\label{eq:vq_loss}
\end{equation}
where $\text{sg}[\cdot]$ is the \textit{stop-gradient} operator. The $\mathcal{L}_{\mathrm{recon}}$ is the reconstruction loss between the input data $\mathbf{x}$ and the reconstructed decoder output $\hat{\mathbf{x}}$. The two $\mathcal{L}_{\mathrm{embed}}$ and $\mathcal{L}_{\mathrm{commit}}$ losses apply only to codebook variables and encoder weight with a weighting hyperparameter $\beta$ to prevent fluctuations from one codebook vector to another. Since the quantization process is non-differentiable, the codebook loss is typically approximated via a straight-through gradient estimator \citep{bengio2013estimating}, such as $\partial \mathcal{L} / \partial f_{\phi}(\mathbf{x}) \approx \partial \mathcal{L} / \partial \mathbf{z}_q(\mathbf{x})$. Both conventional VAE \citep{kingma2013auto} and VQ-VAE \citep{van2017neural} have objective functions consisting of the sum of reconstruction error and latent regularization. To improve performance and convergence rate, an exponential moving average (EMA) update is usually applied for the codebook optimization \citep{van2017neural, razavi2019generating}. Thus, VQ-VAEs serve as a foundation for many advanced generative models, forming the core approach to discrete latent representation.

\subsection{Sequence-to-Sequence Learning}

The Seq2Seq \citep{sutskever2014sequence} model is widely used in sequence prediction tasks such as language modeling and machine translation \citep{dai2015semi, DBLP:journals/corr/LuongLSVK15, DBLP:journals/corr/RanzatoCAZ15}. The model employs an initial LSTM, called the encoder, to process the input sequence $x_{1:N}$ sequentially and produce a substantial fixed-dimensional vector representation, called the context vector. The output sequence $y_{1:T}$ is then derived by a further LSTM, the decoder. A Seq2Seq with parameters $\psi$ estimates the distribution of input sequence $y_{1:M}$ by decomposing it into an ordered product of conditional probabilities:
\begin{equation}
p\big(y_{1:T} | x_{1:N} ; \psi \big) = \prod_{t=1}^{T} p\big(y_t | y_{1:t-1}, x_{1:N} ; \psi \big)
\label{eq:seq2seq}
\end{equation}
During training, the Seq2Seq model typically uses \emph{teacher-forcing} \citep{williams1989learning}, where the target sequence is provided to the decoder at each time step, instead of the decoder using its own previous output as input. This method helps the model converge faster by providing the correct context during training.

\section{Methods} \label{sec:methods}
 
Although VQ-based generative models have demonstrated strong performance in various domains, their fixed codebook size often limits their adaptability to varying datasets or resource constraints. For example, different tasks may require drastically different codebook sizes $K$, ranging from as many as 16{,}384 \citep{razavi2019generating} to as few as 512 \citep{esser2021taming}, requiring developers to either retrain or maintain multiple VQ models at different rates. To address these challenges, we introduce the RAQ framework, which adjusts the VQ model's rate by increasing or decreasing the codebook size $K$. Our method is based on a codebook mapping $\Psi: (\mathbb{R}^d)^{\times K}\longrightarrow (\mathbb{R}^d)^{\times \widetilde{K}}$ for any integer $\widetilde{K} \in \mathbb{N}$, where $\widetilde{K}<K$ or $\widetilde{K}>K$ is possible. In the following sections, we detail our RAQ framework, starting with a Seq2Seq-based multi-rate codebook adaptation strategy, followed by a model-based approach as a simpler alternative.

\subsection{Rate-Adaptive Quantization} \label{sec:raq}

\begin{figure*}[tb]
\begin{center}
\centerline{\includegraphics[width=0.8\linewidth]{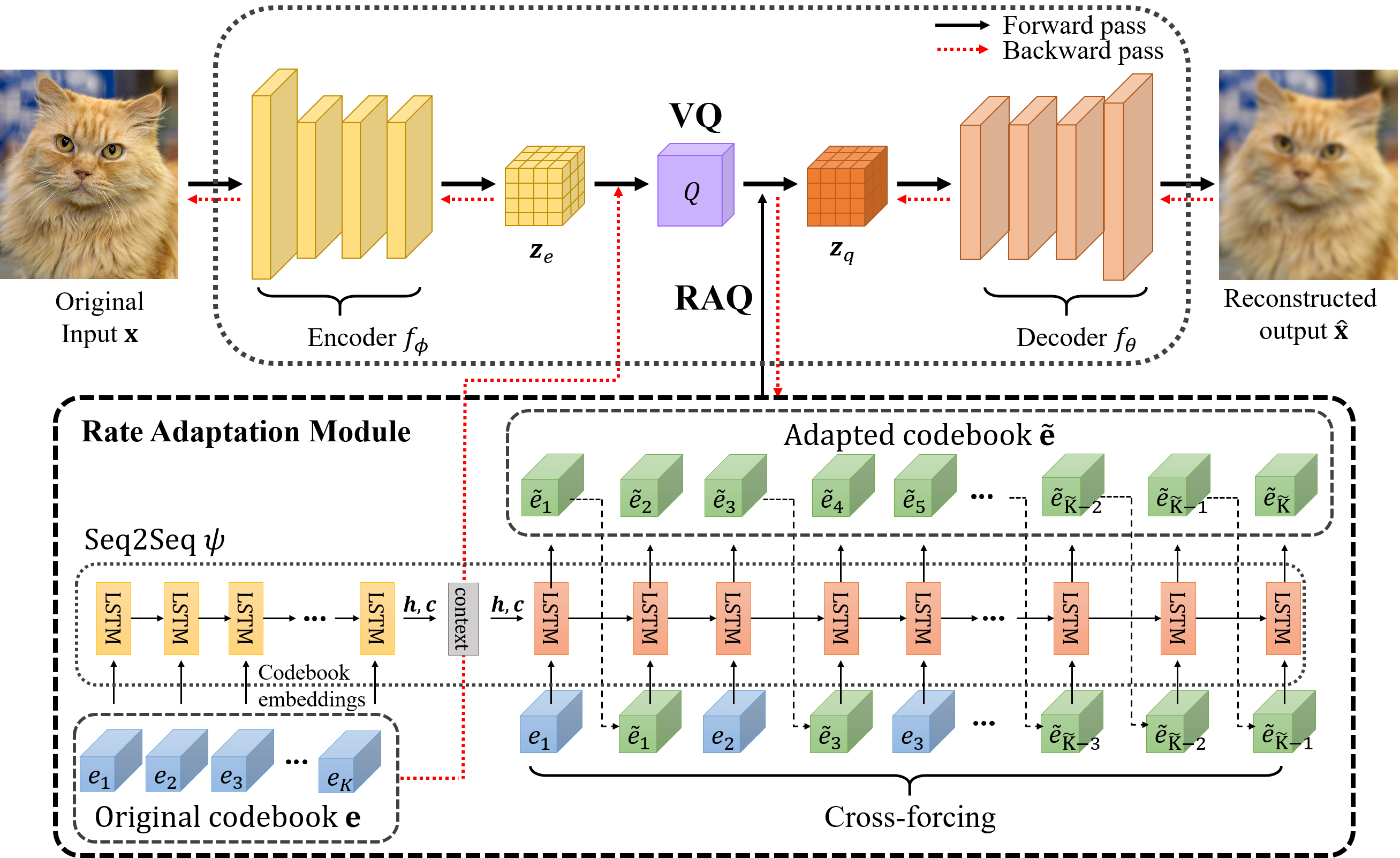}}
\caption{An overview of our RAQ framework applied to a baseline VQ-VAE architecture. During training, the rate adaptation module $G_{\psi}$ employs cross-forcing to generate adapted codebooks $\mathbf{\tilde{e}}$ of arbitrary sizes $\widetilde{K}$ from the original codebook $\mathbf{e}$. At inference, a user-specified $\widetilde{K}$ produces the corresponding adapted codebook for quantization.}
\label{fig:1}
\end{center}
\vskip -0.2in
\end{figure*}

\paragraph{Overview}
The RAQ framework is designed to integrate seamlessly with existing VQ-based generative models without requiring significant architectural modifications. As illustrated in Figure~\ref{fig:1}, RAQ is integrated into a baseline VQ-VAE which consists of an encoder-decoder pair and a trainable original codebook $\mathbf{e}$. The adapted codebook $\mathbf{\tilde{e}}$ is generated by a Seq2Seq model from the original codebook $\mathbf{e}$. During training, the continuous latent representation $f_{\phi}(x)$ of data $\mathbf{x}$ is hierarchically quantized into $\mathbf{z}_q(\mathbf{x}|\mathbf{e})$ and $\mathbf{z}_q(\mathbf{x}|\mathbf{\tilde{e}})$ with $\mathbf{e}$ and $\mathbf{\tilde{e}}$, respectively.

\paragraph{Autoregressive Codebook Generation}
Our rate adaptation module $G_{\psi}$ maps an original codebook $\mathbf{e}$ to an adapted codebook $\mathbf{\tilde{e}}$ via an autoregressive Seq2Seq model. Each codebook vector $e_i$ of $\mathbf{e}$ is treated analogously to a token in language modeling. We learn the Seq2Seq model to output $\widetilde{K}$ adapted codebook vectors. This autoregressive generation ensures that each adapted codebook vector $\tilde{e}_i$ is conditioned on the previously generated ones, maintaining coherence and structural integrity. The adapted codebook $\mathbf{\tilde{e}}$ autoregressively generated as:
\begin{equation}
p\big(\mathbf{\tilde{e}} | \mathbf{e} ; \psi \big) = \prod_{i=1}^{\widetilde{K}} p\big(\tilde{e}_i | \tilde{e}_{<i}, e_{1:K} ; \psi \big)
\label{eq:seq2seq_raq}
\end{equation}
where $\tilde{e}_{<i}$ denotes the sequence of adapted codebook vectors generated before step $i$. In typical Seq2Seq tasks, models are trained using a negative log-likelihood loss that emphasizes the prediction of the next token in a sequence. In our RAQ framework, the order of the adapted codebook vectors $\tilde{e}_{i:\widetilde{K}}$ does not hold inherent meaning; rather, it is the overall distribution of these embeddings that is important. RAQ directly incorporates the loss function of the baseline VQ model into the training of the Seq2Seq model. This integration ensures that the generation of adapted codebooks maintains the desired distributional properties essential for high-quality reconstruction without imposing an arbitrary sequence order or requiring additional loss terms.

\paragraph{Codebook Encoding}
The first step is initializing the target codebook size $\widetilde{K}$. The RAQ-based model is trained with arbitrary codebook sizes $\widetilde{K}$ during training. In the test phase, the Seq2Seq model generates the adapted codebook $\mathbf{\tilde{e}}$ at the desired rate specified by the user. Each original codebook vector $e_i$ is sequentially processed by the LSTM cells, where the hidden and cell states ($\boldsymbol{h}, \boldsymbol{c}$) capture the contextual information of each original codebook vector. This encoding captures dependencies and contextual relationships within the original codebook, providing a foundation for generating a coherent adapted codebook.

\paragraph{Codebook Decoding via Cross-Forcing}
We introduce a novel cross-forcing strategy to effectively train the Seq2Seq model for generating variable-rate codebooks. Traditional teacher-forcing training strategy \citep{williams1989learning} may not be suitable when the target adapted codebook $\mathbf{\tilde{e}}$ consists of sequences that are much longer than the original codebook. Cross-forcing mitigates this by alternating between teacher-forcing and free-running mode in professor-forcing \citep{lamb2016professor}.

During training, the cross-forcing strategy operates as:

\begin{itemize}
    \item \textbf{Teacher-Forcing Phase.}
    For odd indices $i$ up to $2K$ (i.e., $1 \leq i \leq 2K$ and $i$ is odd), the model uses the corresponding original codebook vector $e_j$ as input, where $j = \frac{i+1}{2}$:
    $$
    \tilde{e}_i = \text{LSTM}_{\psi}(\tilde{e}_{<i}, e_j, \boldsymbol{h}, \boldsymbol{c}).
    $$
    This ensures that the fundamental distributional features of the original codebook are preserved during the early generation steps.
    \item \textbf{Free-Running Phase.}
    For even indices $i$ up to $2K$ ($i$ is even and $i \leq 2K$), and for all indices beyond $2K$ (i.e., $i > 2K$), the model relies on its previously generated adapted codebook vector $\tilde{e}_{i-1}$:
    $$
    \tilde{e}_i = \text{LSTM}_{\psi}(\tilde{e}_{<i}, \tilde{e}_{i-1}, \boldsymbol{h}, \boldsymbol{c}).
    $$
    By switching to its own outputs, the model learns to maintain coherence and consistency across the adapted codebook vectors for different sizes of $\widetilde{K}$.
\end{itemize}

Training the codebook via cross-forcing is a key component of our RAQ approach. We provide an empirical evaluation of its effectiveness in Appendix~\ref{appendix_cross_forcing}.

\paragraph{Training Procedure}

\begin{figure}[tb]
    \centering
    \begin{minipage}[c]{0.6\linewidth}
\begin{algorithm}[H]
    \centering
    \caption{Training procedure of RAQ}
    \label{alg:training_raq}
    \begin{algorithmic}
        \State {\bfseries Input:} $\mathbf{x}$ (batch of training data)
        \For{$\mathbf{x} \in$ train dataset $\mathcal{D}$} 
            \State \hspace{-1em} $\triangleright$ \ Quantize encoder output $f_{\phi}(\mathbf{x})$ with $\mathbf{e}$.
            \State $\mathbf{z}_q(\mathbf{x}|\mathbf{e}) \gets Q\left(f_{\phi}(\mathbf{x})|\mathbf{e}\right)$
            \State \hspace{-1em} $\triangleright$ \ Generate $\mathbf{\tilde{e}}$ from Rate adaptation module $G_{\psi}$.
            \State $\mathbf{\tilde{e}} \gets G_{\psi}(\mathbf{e})$ 
            \State \hspace{-1em} $\triangleright$ \ Quantize encoder output $f_{\phi}(\mathbf{x})$ with $\mathbf{\tilde{e}}$.
            \State $\mathbf{z}_q(\mathbf{x}|\mathbf{\tilde{e}}) \gets Q\left(f_{\phi}(\mathbf{x})|\mathbf{\tilde{e}}\right)$
            \State $\hat{\mathbf{x}}_{\mathbf{e}}, \hat{\mathbf{x}}_{\mathbf{\tilde{e}}} \gets f_{\theta}(\mathbf{z}_q(\mathbf{x}|\mathbf{e})), f_{\theta}(\mathbf{z}_q(\mathbf{x}|\mathbf{\tilde{e}}))$
            \State \hspace{-1em} $\triangleright$ Compute losses:
            \State $\mathcal{L}_{\text{VQ}} \gets \text{Compute VQ Loss}$ \ (Eq.~\eqref{eq:vq_loss})
            \State $\mathcal{L}_{\text{RAQ}} \gets \text{Compute RAQ Loss}$ \ (Eq.~\eqref{eq:raq_loss})
            \State \hspace{-1em} $\triangleright$ Update model parameters:
            \State $\phi, \theta, \mathbf{e} \gets \text{Update}(\mathcal{L_\text{VQ}})$
            \State $\phi, \theta, \psi, \mathbf{e} \gets \text{Update}(\mathcal{L_\text{RAQ}})$
        \EndFor
    \end{algorithmic}
\end{algorithm}
\end{minipage}
\end{figure}

The Seq2Seq parameters $\psi$ are optimized for the overall reconstruction and codebook utilization over variable codebook sizes. We define the RAQ loss $\mathcal{L}_\text{RAQ}$ to incorporate both the standard VQ loss \eqref{eq:vq_loss} and the RAQ-driven loss $\mathcal{L}_\text{RAQ}\big(\phi,\theta, \psi, \mathbf{e};\mathbf{x}\big)=$
\begin{equation}
    \log p_{\theta}\big(\mathbf{x}|\mathbf{z}_q(\mathbf{x}|G_{\psi}(\mathbf{e}))\big) + \big|\big|\text{sg}\left[f_{\phi}(\mathbf{x})\right]-\mathbf{z}_q\big(\mathbf{x}|G_{\psi}(\mathbf{e})\big)\big|\big|_2^2 + \beta\big|\big|\text{sg}\left[\mathbf{z}_q\big(\mathbf{x}|G_{\psi}(\mathbf{e})\big)\right]-f_{\phi}\big(\mathbf{x}\big)\big|\big|_2^2.
\label{eq:raq_loss}
\end{equation}

Algorithm~\ref{alg:training_raq} outlines the overall training flow, where codebook adaptation module $G_{\psi}$ operates at every iteration. This allows RAQ to adaptively learn multiple rates without retraining separate models.

\subsection{Model-Based RAQ}\label{sec:model-based_RAQ}
We propose the \emph{model-based} RAQ method as an alternative approach to adapt the codebook rate without requiring retraining the VQ model. In contrast to our Seq2Seq-based RAQ, this approach does not learn new parameters but instead performs clustering on a pre-trained VQ codebook $\mathbf{e}$, dynamically adjusting its size to a target $\widetilde{K}$.
 
 \paragraph{Codebook Clustering}
To produce an adapted codebook $\mathbf{\tilde{e}}$ of size $\widetilde{K}$, we utilize differentiable $k$-means (DKM) \citep{cho2021dkm}, originally introduced for compressing model weights via layer-wise clustering. In our case, we repurpose DKM to cluster the embedding vectors in $\mathbf{e}$, effectively reducing (or expanding) the codebook and preserving essential structure. Details of this clustering process are provided in Appendix~\ref{appendix_clustering}. Additionally, DKM enables inverse functionalization (IKM) to accommodate increases in codebook size, allowing for both rate-reduction and rate-increment tasks.

\paragraph{Reducing the Rate $(\widetilde{K} < K)$}
In the rate-reduction task, DKM performs iterative, differentiable codebook clustering on $\widetilde{K}$ clusters. Let $\mathbf{C}=\{c_j\}_{j=1}^{\widetilde{K}}$ be the cluster centers for the original codebook $\mathbf{e}$. The process is as follows:

\begin{itemize}
    \item 
    Initialize the centroids $\mathbf{C}=\{c_j\}_{j=1}^{\widetilde{K}}$ by randomly selecting $\widetilde{K}$ codebook vectors from $\mathbf{e}$ or by using $k$-means++. The last updated $\mathbf{C}$ is used in subsequent iterations.
    \item 
    Compute the Euclidean distance between each $e_i$ and $c_j$, denoting $D_{i,j} = -f(e_i, c_j)$ to form the matrix $\boldsymbol{D}$.
    \item 
    Form the attention matrix $\boldsymbol{A}$ via a softmax with temperature $\tau$, where each row satisfies $A_{i, j}=\frac{\exp \left(\frac{D_{i,j}}{\tau}\right)}{\sum_k \exp \left(\frac{D_{i,k}}{\tau}\right)}$.
    \item 
    Compute the candidate centroids $\widetilde{\mathbf{C}}=\{\tilde{c}_j\}$ by $\tilde{c}_j=\frac{\sum_i A_{i, j} e_i}{\sum_i A_{i, j}}$, then update $\mathbf{C} \leftarrow \widetilde{\mathbf{C}}$.
    \item 
    Repeat until $\|\mathbf{C}-\widetilde{\mathbf{C}}\|\le \epsilon$ or the iteration limit is reached. We then multiply $\boldsymbol{A}$ by $\mathbf{C}$ to obtain the final $\mathbf{\tilde{e}}$.
\end{itemize}

The above iterative process can be summarized as follows:
\begin{equation}
\label{obj_dkm}
\mathbf{\tilde{e}} = \underset{\mathbf{\tilde{e}}}{\arg\min} \ \mathcal{L}_{\text{DKM}}(\mathbf{e}; \mathbf{\tilde{e}})= \underset{\mathbf{C}}{\arg\min} \ |\mathbf{C} - \boldsymbol{A}\mathbf{C}| = \underset{\mathbf{C}}{\arg\min} \sum_{j=1}^{\widetilde{K}} \left|c_j-\frac{\sum_i A_{i, j} e_i}{\sum_i A_{i, j}}\right|.
\end{equation}
In \citep{cho2021dkm}, the authors introduced DKM for soft-weighted clustering in model compression, allowing one to enforce different levels of hardness. After convergence, we assign each codebook vector to its nearest centroid according to the last attention matrix $\boldsymbol{A}$, thereby finalizing the compressed codebook.

\paragraph{Increasing the Rate $(\widetilde{K} > K)$}
To handle scenarios where a larger codebook is beneficial, we adopt an inverse functional DKM (IKM) method. This approach allows new codebooks to be added by approximating the posterior distribution of an existing codebook. While $k$-means clustering is effective for compressing codebook vectors, it has algorithmic limitations that prevent the augmentation of additional codebooks. To address this, we introduce the Inverse functional DKM (IKM), a technique for increasing the number of codebooks. This iterative method aims to approximate the posterior distribution of an existing generated codebook. We use maximum mean discrepancy (MMD) to compare the distribution difference between the base codebook and the clustered generated codebook, where MMD is a kernel-based statistical test technique that measures the similarity between two distributions \citep{gretton2012kernel}.

Assuming the original codebook vector $\mathbf{e}$ of size $K$ already trained in the baseline VQ model, the process of generating the codebook $\mathbf{\tilde{e}}$ using the IKM algorithm is performed as follows:

\begin{itemize}
\item Initialize a $d$-dimensional adapted codebook vector $\mathbf{\tilde{e}}=\left\{\tilde{e}_i\right\}_{i=1}^{\widetilde{K}}$ as $\mathbf{\tilde{e}} \sim \mathcal{N}(0, d^{-\frac{1}{2}} \boldsymbol{I}_{\widetilde{K}})$

\item Cluster $\mathbf{\tilde{e}}$ via the DKM process (Eq~\eqref{obj_dkm}): $g_{\text{DKM}}(\mathbf{\tilde{e}}) = \underset{g_{\text{DKM}}(\mathbf{\tilde{e}}) }{\arg\min} \ \mathcal{L}_{\text{DKM}}(\mathbf{\tilde{e}}; g_{\text{DKM}}(\mathbf{\tilde{e}}))$.

\item Calculate the MMD between the true original codebook $\mathbf{e}$ and the DKM clustered $g_{\text{DKM}}(\mathbf{\tilde{e}})$.

\item Optimize $\mathbf{\tilde{e}}$ to minimize the MMD objective $\mathcal{L}_{\text{IKM}}(\mathbf{e}; \mathbf{\tilde{e}})=\mathrm{MMD}(\mathbf{e}, g_{\text{DKM}}(\mathbf{\tilde{e}})) + \lambda ||\mathbf{\tilde{e}}||^2$.
\end{itemize}

where $\lambda$ is the regularization parameter controlling the strength of the L2 regularization term. The IKM process can be summarized as $\mathbf{\tilde{e}} = \underset{\mathbf{\tilde{e}}}{\arg\min} \ \mathcal{L}_{\text{IKM}}(\mathbf{e}; \mathbf{\tilde{e}})$. Since DKM does not block gradient flow, we easily can update the codebook $\mathbf{\tilde{e}}$ using stochastic gradient descent (SGD) as $\mathbf{\tilde{e}}=\mathbf{\tilde{e}}-\eta \nabla \mathcal{L}_{\text{IKM}}(\mathbf{e}, \mathbf{\tilde{e}})$.

Under our model-based RAQ strategy, the clustered codebook $\mathbf{\tilde{e}}$ can quantize the encoded vectors $\mathbf{z}_q(\mathbf{x}|\mathbf{\tilde{e}})$ at different rates, without adding parameters to the original VQ model. Because DKM retains gradient flow, one can adjust cluster assignments seamlessly, either offline or online. During offline training, the model can adopt clusters best suited for minimizing the VQ loss. Although we do not explore a multi-codebook scenario here, it would be straightforward to extend DKM for that purpose by tuning $\widetilde{K}$ during training and hierarchically optimizing the multi-codebook clusters with the model.

\section{Related Work}

\paragraph{VQ and its Improvements}
The VQ-VAE \citep{van2017neural} has inspired numerous developments since its inception. \citet{lancucki2020robust, williams2020hierarchical, zheng2023online} proposed codebook reset or online clustering methods to mitigate \emph{codebook collapse} \citep{takida2022sq}, improving training efficiency. \citet{tjandra2019vqvae} introduced a conditional VQ-VAE that generates magnitude spectrograms for target speech using a multi-scale codebook-to-spectrogram inverter from the VQ codebook. SQ-VAE \citep{takida2022sq} incorporated stochastic quantization and a trainable posterior categorical distribution to enhance VQ performance. Based on SQ-VAE, \citet{vuong2023vector} proposed VQ-WAE, leveraging Wasserstein distance to encourage uniform discrete representations. Several works have introduced substantial structural changes to VQ modeling; for instance, \citet{lee2022autoregressive} employed a two-step residual quantization framework for high-resolution images, while \citet{mentzer2023finite} replaced VQ with Finite Scalar Quantization (FSQ) to address codebook collapse. By contrast, instead of substantial structural changes, our approach focuses on making rate-adaptive VQ without substantially altering the quantization mechanism or architecture, allowing it to scale effectively in both basic and advanced VQ models.

\paragraph{Variable-Rate Neural Image Compression}
Several studies have proposed variable-rate image compression based on autoencoders and VAEs \citep{yang2020variable, choi2019variable, cui2020g}, or using recurrent neural networks \citep{johnston2018improved}. \citet{song2021variable} introduced spatial feature transforms for compression, while \citet{duong2023multi} combined learned transforms and entropy coding in a single model aligned with the rate-distortion curve. Although these methods successfully accommodate multiple rates, most rely on continuous latent representations and do not directly address VQ-based modeling. In particular, variable-rate compression grounded in VQ remains underexplored, since VQ enforces a discrete codebook to maximize resource utilization. Adapting this discrete representation for flexible rates therefore requires carefully balancing codebook size and quantization efficiency. Our RAQ framework specifically tackles this gap by enabling a single VQ-based model to accommodate various codebook sizes without retraining.

\section{Experiments}

\subsection{Experimental Setup}

\paragraph{Baseline VQ Models}
To demonstrate RAQ's flexibility in covering multiple rates within a single architecture, we adapt the conventional VQ-VAE \citep{van2017neural}, the two-level hierarchical VQ-VAE (VQ-VAE-2) \citep{razavi2019generating}, and the stage-1 VQGAN model \citep{esser2021taming} as baseline VQ models.

\paragraph{Settings}
We perform empirical evaluations on 3 vision datasets: CIFAR10 ($32\times32$) \citep{krizhevsky2009learning} and CelebA ($64\times64, 128\times128$) \citep{liu2015faceattributes}, and ImageNet ($256\times256$) \citep{ILSVRC15}. We use identical architectures and parameters for all comparison methods. The codebook embedding dimension $d$ is set to 64 for CIFAR10 and CelebA, and to 128 for ImageNet. The adapted codebook sizes range from 16 to 1{,}024 for CIFAR10, 32 to 2{,}048 for CelebA, and 32 to 4{,}096 for ImageNet, while the original codebooks of baseline VQ models are fixed. RAQ-based models set the original codebook size to the middle of the range. We also provide details on each model’s parameter count and complexity in Appendix~\ref{appendix_model_complexity}.

\paragraph{Evaluation Metrics}
We quantitatively evaluated our method using Peak-Signal-to-Noise-Ratio (PSNR), Structural Similarity Index Measure (SSIM), reconstructed Fréchet Inception Distance (rFID) \citep{heusel2017gans}, and codebook perplexity. PSNR measures the ratio between the maximum possible power of a signal and the power of the corrupted noise affecting data fidelity \citep{6263880}. SSIM assesses structural similarity between two images \citep{wang2004image}. rFID assesses the quality of reconstructed images by comparing the distribution of features extracted from the test data with that of the original data. Codebook perplexity, defined as \( e^{-\sum^{\widetilde{K}}_{i}p_{e_{i}} \log p_{e_{i}}} \) where \( p_{e_{i}} = \frac{N_{e_i}}{\sum^{\widetilde{K}}_{j}N_{e_j}} \) and \( N_{e_i} \) represents the encoded number for latent representation with codebook \( e_i \), indicates a uniform prior distribution when the perplexity value reaches the codebook size $\widetilde{K}$.

\begin{figure*}[tb]
    \centering
    \begin{subfigure}[b]{0.95\linewidth}
        \centering
        \includegraphics[width=\linewidth]{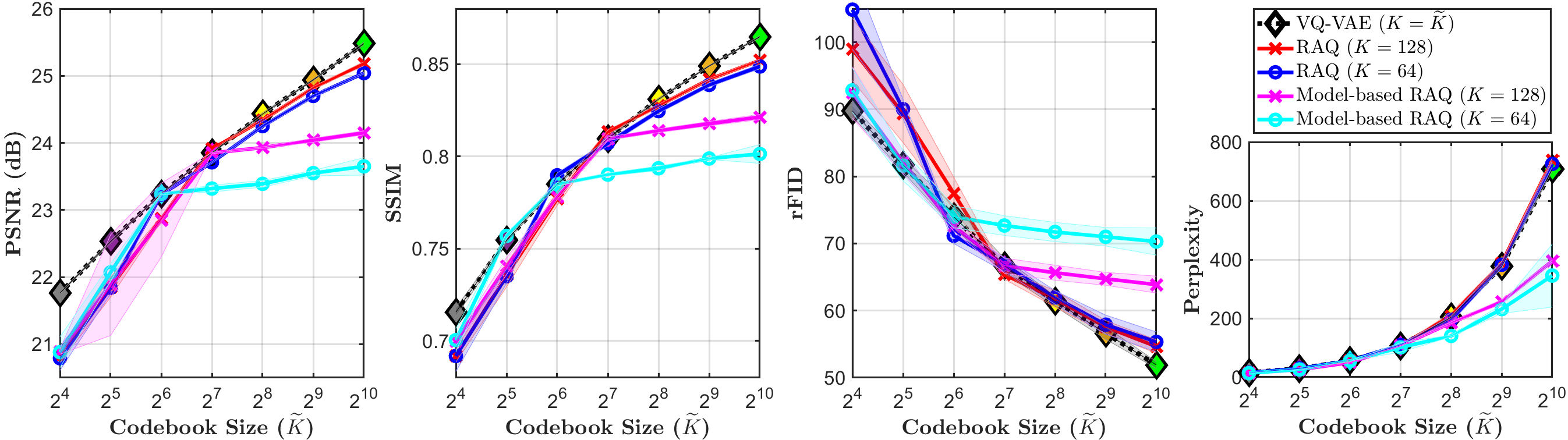}
        \caption{CIFAR10, baseline model: VQ-VAE \citep{van2017neural}}
        \label{cifar10_vqvae}
    \end{subfigure}
    \begin{subfigure}[b]{0.95\linewidth}
        \centering
        \includegraphics[width=\linewidth]{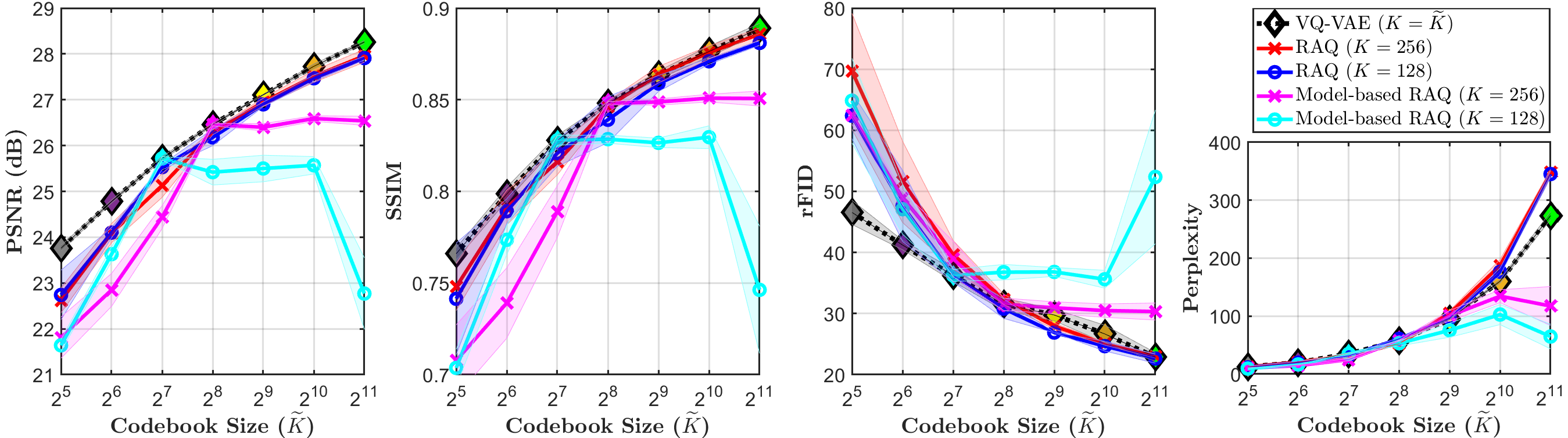}
        \caption{CelebA, baseline model: VQ-VAE \citep{van2017neural}}
        \label{celeba_vqvae}
    \end{subfigure}
    \caption{\textbf{Reconstruction Performance} on (a) CIFAR-10 and (b) CelebA at various codebook sizes $\widetilde{K}$. Higher values are better for PSNR, SSIM, and codebook perplexity, while lower values are better for rFID. \emph{Black lines}: individual VQ-VAE models for each codebook size $\widetilde{K}$. \emph{Colored lines}: single RAQ-based model adapted from codebook size $K$ to $\widetilde{K}$. The shaded area indicates the 95.45\% confidence interval based on 4 runs with different seeds.}
    \label{fig:result_quant}
    \vskip -0.2in
\end{figure*}

\subsection{Quantitative Evaluation} \label{quant_eval}

We empirically evaluate the effectiveness of RAQ using the baseline VQ-VAE \citep{van2017neural} on CIFAR10 ($32\times32$) and CelebA ($64\times64$) for image reconstruction. To establish a robust baseline, we trained multiple VQ-VAE models with varying codebook sizes $K$ as fixed-rate benchmarks. We then tested RAQ’s adaptability by dynamically adjusting the codebook size $\widetilde{K}$ within a single VQ-VAE. Figure~\ref{fig:result_quant} shows the comparative results.

RAQ closely matches the performance of multiple fixed-rate VQ-VAE models across most metrics, demonstrating its ability to maintain high reconstruction quality while offering rate flexibility. Specifically, under identical compression rates and network architectures, all RAQ variants achieve PSNR and SSIM scores nearly on par with their fixed-rate counterparts (e.g., within about 0.94 dB difference in PSNR on CIFAR10). This indicates that RAQ preserves essential image details and structural integrity 
despite dynamically adjusting the codebook size.

When we increase the rate (i.e., expand $\widetilde{K}$), RAQ occasionally shows slightly lower PSNR and SSIM but improves rFID, reflecting improved perceptual quality and better alignment with the dataset distribution. For instance, at $\widetilde{K} = 512$ on CelebA, rFID improves by up to about 9.6\% compared to a fixed-rate VQ-VAE with the same codebook size, highlighting RAQ’s ability to maintain visual coherence and realism in generative tasks. Additionally, RAQ exhibits higher perplexity at larger $\widetilde{K}$ on CelebA, indicating it captures more diverse representations. By adaptively adjusting the codebook size, RAQ can better align the discrete latent space with data complexity.

However, the model-based RAQ variant typically underperforms our Seq2Seq-based approach except in certain rate-reduction tasks. For example, at $\widetilde{K} = 128$ on CIFAR10, model-based RAQ achieves around 1.1 dB higher PSNR than the Seq2Seq-based RAQ. Although model-based RAQ is less consistent overall, it can be advantageous where retraining is infeasible or when lower compression rates suffice.
                  
Overall, RAQ offers substantial advantages in portability and reduced complexity. By consolidating multiple fixed-rate VQ models into a single adaptable framework, RAQ saves training and storage overhead while simplifying deployment. Although minor performance tradeoffs may arise at certain rates, the combination of flexibility and efficiency makes RAQ  attractive for a variety of generative tasks. 

\begin{figure}[tb]
\begin{center}
\centerline{\includegraphics[width=0.85\textwidth]{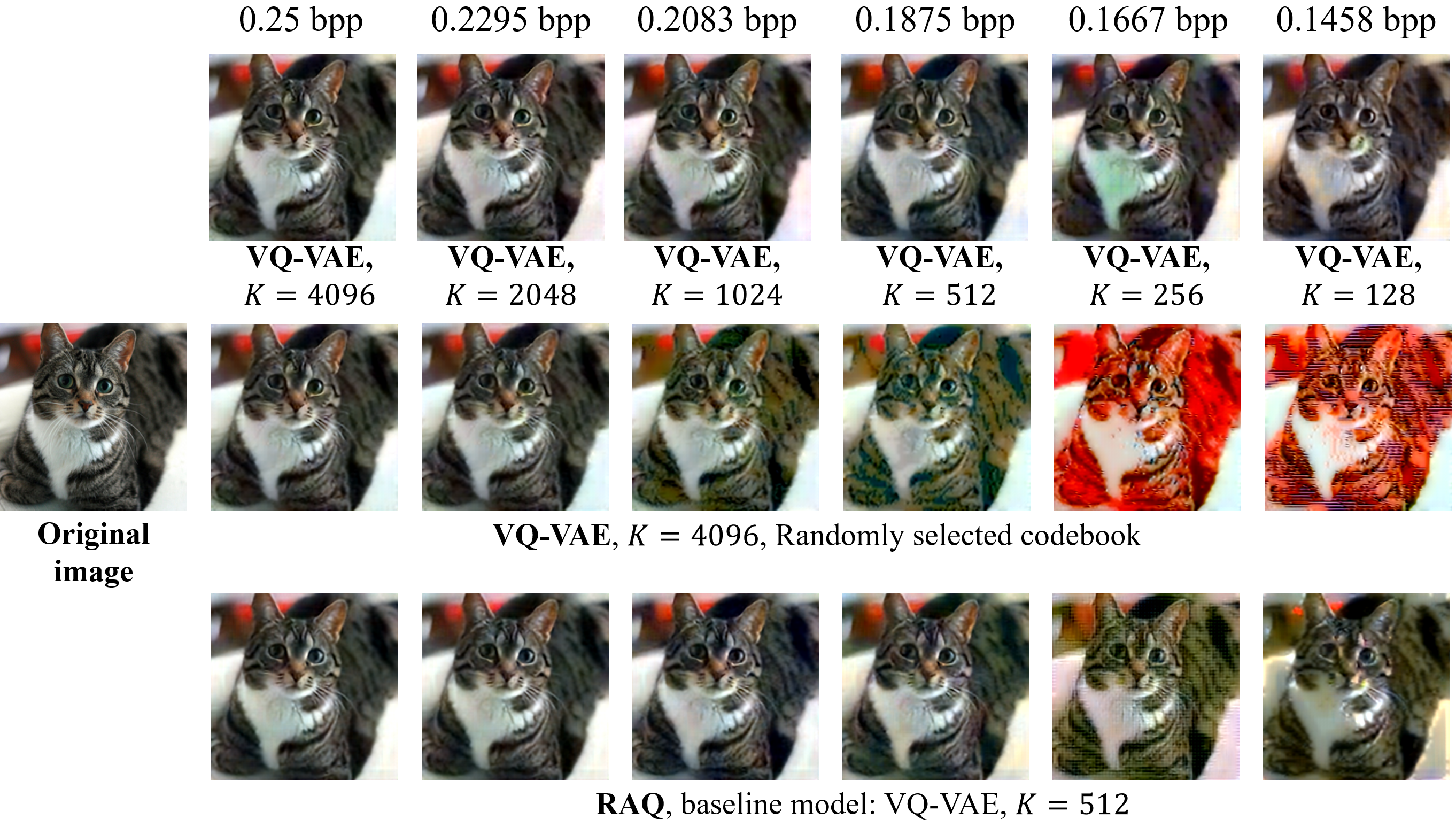}}
\caption{\textbf{Qualitative comparison} on ImageNet ($256\times256$) at different compression rates. \textbf{Top row}: Fixed-rate VQ-VAEs trained separately at each rate. \textbf{Middle row}: A single VQ-VAE ($K=4096$) with randomly selected codebooks. \textbf{Bottom row}: Our RAQ with VQ-VAE ($K=512$) with adapting the codebook size.}
\label{fig:result_quali}
\end{center}
\vskip -0.2in
\end{figure}

\subsection{Qualitative Evaluation} \label{quali_eval}

For our qualitative evaluation, We compare a single RAQ-based model against multiple VQ-VAEs trained at different rates (0.1458 bpp to 0.25 bpp) on ImageNet ($256 \times 256$). As illustrated in Figure~\ref{fig:result_quali}, each fixed-rate VQ-VAE (top row) shows a progressive decline in image quality as the rate decreases, consistent with the quantitative evaluation. Unlike RAQ-based reconstruction, randomly selecting codebooks from a single VQ-VAE trained at $K=4096$ (middle row) results in color distortions and inconsistent hues, especially at 0.1667 bpp. Despite retaining the basic structure, the mismatched usage of codebooks still produces unnatural appearances. By contrast, our RAQ-based VQ-VAE (bottom row), trained at a low-rate base codebook of 0.1875 bpp (roughly $K=512$), effectively preserves high-level semantic features and color fidelity using only a single model. Notably, it recovers finer details (e.g., the cat’s whiskers) far better than models relying on randomly selected codebooks. Although image quality declines slightly at the lowest bpp, largely due to the limited capacity of the baseline VQ-VAE, this issue can be mitigated by using more advanced VQ architectures or refining training procedures. Training RAQ with a smaller original codebook size $K$ can also help reduce performance degradation at lower rates.

For further demonstrations, we refer readers to Figure~\ref{fig:result_quali_kodak} in Appendix~\ref{appendix_qualitative_results}, where our RAQ-based VQ-VAE-2 maintains high-quality reconstructions across a range of codebook sizes and higher-resolution images. In future work, we plan to integrate RAQ with autoregressive generative models (e.g., PixelCNN \citep{pmlr-v48-oord16} or PixelSNAIL \citep{pmlr-v80-chen18h}) to further enhance reconstruction (or generation) fidelity. 

\subsection{Broader Applications and Discussion}\label{sec:extended_applicability}

\afterpage{
\begin{table*}[tb]
\caption{Comparison of \emph{MS-SSIM} and \emph{perplexity} scores for two advanced VQ-based generative models. Symbols: $\dagger$ denotes a single model using randomly selected $\widetilde{K}$ codebooks for reconstructions; * indicates models trained with \textbf{VQ-VAE-2} on CelebA ($128\times128$); ** indicates models trained with \textbf{stage-1 VQGAN} on ImageNet ($256\times256$); $\S$ and $\S\S$ represent results from the same model at corresponding rates.
}
\label{table:result_applicability}
\vskip -0.2in
\begin{center}
\resizebox{\textwidth}{!}{
{\setlength{\tabcolsep}{2pt}
\begin{tabular}{@{} l c cccccc @{}}
\toprule                                
( \emph{MS-SSIM} $\uparrow$ \ / \ \emph{Perplexity} $\uparrow$ )& \multirow{2}{*}{$K$}    & \multicolumn{6}{c}{\textbf{Adapted Codebook Size} $\widetilde{K}$}  \\ \cmidrule(lr){3-8}
\textbf{Method}                                 &                           & 2048                              & 1024                                    & 512                                   & 256                                   & 128                                   & 64                    \\
\midrule
VQ-VAE-2 \citep{razavi2019generating}           & $\widetilde{K}$           & $\mathbf{0.9884}^\S$ / $334.7^\S$ & $\mathbf{0.9865}$ / 183.2               & 0.9842  / 103.1                       & $\mathbf{0.9810}$ / 61.6              & $\mathbf{0.9780}$  / 36.3             & $\mathbf{0.9712}$  / 21.1    \\
VQ-VAE-2$^\dag$ \citep{razavi2019generating}    & 2048                      & $\mathbf{0.9884}^\S$ / $334.7^\S$ & 0.9717 / 178.1                          & 0.9642  / 103.0                       & 0.9514 / 60.4                         & 0.7266  / 14.1                        & 0.6094  / 6.9     \\
Model-based RAQ* (Ours)                         & 2048                      & $\mathbf{0.9884}^\S$ / $334.7^\S$ & 0.9813 / 131.1                          & 0.9733  / 53.8                        & 0.9658 / 30.6                         & 0.9518  / 15.6                        & 0.9320  / 10.6    \\   
\textbf{RAQ* (Ours)}                            & $\mathbf{256}$            & 0.0.9881 / $\mathbf{465.2}$       & $\mathbf{0.9865}$ / $\mathbf{239.7}$    & $\mathbf{0.9847}$ / $\mathbf{133.6}$  & $\mathbf{0.9809}$ / $\mathbf{67.03}$  & 0.9771 / $\mathbf{38.7}$              & 0.9693 / $\mathbf{24.2}$ \\
\midrule
                                                &                           & 512                                       & 256                                       &   128                                     & 64                                        & 32                     & -      \\
\midrule
VQGAN \citep{esser2021taming}                   & $\widetilde{K}$           & $\mathbf{0.7543}^\S$  /  $296.3^\S\S$     & 0.7383            / 140.0                 & 0.7221            / 75.0                  & 0.7015 / 40.1                             & $\mathbf{0.6834}$ / 21.87           & - \\
VQGAN$^\dag$ \citep{esser2021taming}            & 512                       & $\mathbf{0.7543}^\S$  /  $296.3^\S\S$     & 0.7311            / 148.5                 & 0.7035            / 74.1                  & 0.6614 / 37.2                             & 0.5801 / 18.89              & - \\
Model-based RAQ** (Ours)                        & 512                       & $\mathbf{0.7543}^\S$  /  $296.3^\S\S$     & 0.7381            / 147.1                 & 0.7182            / 75.3                  & 0.6908 / 39.6                             & 0.6546 / 21.00            & - \\
\textbf{RAQ** (Ours)}                           & 64                        & $0.7496$ /  $\mathbf{353.9}$              & $\mathbf{0.7409}$ / $\mathbf{181.3}$        & $\mathbf{0.7265}$ / $\mathbf{86.1}$       & $\mathbf{0.7138}$ / $\textbf{46.8}$        & 0.6793 / $\mathbf{23.6}$                 & - \\                           
\bottomrule
\end{tabular}
}
}
\end{center}
\vskip -0.1in
\end{table*}
}

\paragraph{Applicability}
We extend RAQ to two advanced VQ-based generative models: \emph{VQ-VAE-2} \citep{razavi2019generating} and \emph{stage-1 VQGAN} \citep{esser2021taming}. VQ-VAE-2 adopts a hierarchical structure for richer representation, while VQGAN employs adversarial and perceptual objectives for enhanced image fidelity. Table~\ref{table:result_applicability} summarizes MS-SSIM and perplexity results on CelebA ($128\times128$) and ImageNet ($256\times256$). As noted in Figure~\ref{fig:result_quali}, training with a large codebook and then randomly selecting the codebook (second row) leads to significant degradation, especially when more than half of the codebook is removed. By contrast, our RAQ-based models (fourth rows) preserve or even surpass the quality of individually trained baselines over various codebook sizes, adapting the codebook without harming fidelity. These results confirm that RAQ generalizes seamlessly from vanilla VQ-VAE to more complex hierarchical (VQ-VAE-2) and GAN-based (VQGAN) architectures.

\paragraph{Codebook Usability}
Following previous observations \citep{wu2020vector, takida2022sq, vuong2023vector}, we note that increasing the codebook size tends to raise the codebook perplexity, often leading to better reconstruction. In Table~\ref{table:result_applicability}, the bottom-level latents of \emph{VQ-VAE-2} show notably low perplexities (around 16\% to 33\% of capacity), reflecting the hierarchical design of top- and bottom-level codes. In contrast, our RAQ approach achieves higher perplexities (22\% to 38\% of capacity), indicating more efficient usage of the codebook. A similar trend emerges for \emph{stage-1 VQGAN} suggesting that even adversarially trained generative models benefit from RAQ’s flexible codebook adaptation. This improvement is especially evident at larger codebook sizes, where RAQ’s autoregressive generation aligns more effectively with the data distribution and balances code usage across embeddings. Such behavior aligns with the broader objective of maximizing entropy in discrete representations, implying that RAQ can unlock more latent capacity otherwise underutilized in hierarchical VQ setups.

\paragraph{Model-Based RAQ}
Although model-based RAQ allows rate adaptation without retraining, it leverages DKM and IKM for codebook increasing and decreasing tasks, making it sensitive to initialization (particularly for $\widetilde{K} > K$). Poor initialization can lead to suboptimal local minima, causing newly formed codebooks to be placed ineffectively in the embedding space, and thus underperform compared to our main RAQ framework. As shown in Table~\ref{table:result_applicability}, while model-based RAQ can slightly boost the baseline in rate-reduction tasks, it still shows noticeable performance limitations overall. Nevertheless, this approach may be attractive for practitioners with limited computing resources, as it only requires loading and clustering the original codebook embeddings. Larger architectures, such as ViT-VQGAN \citep{yuvector}, might benefit further from this strategy, given the greater overhead associated with retraining complex models.

\paragraph{Additional Ablation}
We present an ablation study on our cross-forcing strategy in Appendix~\ref{appendix_cross_forcing}, demonstrating its effectiveness in stabilizing codebook generation for larger codebooks (\(\widetilde{K} > K\)). As shown in Table~\ref{result_cross_forcing}, applying cross-forcing can yield up to a 4.9\% improvement in rFID and noticeable gains in PSNR and SSIM compared to a non-cross-forced variant. However, when \(\widetilde{K} \le K\), cross-forcing sometimes underperforms slightly, likely due to its design focus on scaling codebooks beyond the original size. Overall, these results confirm that cross-forcing helps RAQ reliably expand the codebook while incurring only minor tradeoffs at smaller rates.

\paragraph{Computational Cost}
For computational overhead, we refer the readers to Appendix~\ref{appendix_time}. Although RAQ adds a Seq2Seq module that increases per-epoch training time relative to a single fixed-rate VQ, its overall cost remains practical when considering scenarios requiring multiple fixed-rate models. During inference, our current implementation re-generates the adapted codebook for each mini-batch, leading to significant overheads in raw numbers. However, this can be greatly reduced by precomputing the adapted codebook once and reusing it for subsequent samples in real-world streaming or deployment scenarios.

\section{Conclusion}
We proposed Rate-Adaptive Quantization (RAQ) for VQ-based generative models, a method that enables dynamic compression rate adjustment without retraining. This is especially relevant for real-world scenarios with varying bandwidth or computational constraints. Our results show that a single RAQ-based model can cover multiple compression levels with minimal performance loss, eliminating the need for multiple fixed-rate models. We also investigated a simpler model-based RAQ variant that, despite certain limitations, provides a convenient post-hoc alternative when retraining is not feasible. As it only requires adapting codebook embeddings, it remains feasible even for those with limited computing resources. RAQ integrates easily into standard VQ-based architectures, requiring minimal changes. We envision RAQ as a practical and flexible tool for dynamic rate control across diverse VQ tasks, driving advances in both the theoretical aspects of discrete representation learning and the practical fields of generative modeling.

\section*{Impact Statement}

RAQ is a rate-adaptive extension of VQ that can be applied in all domains where VQ methods are used, including image, audio, and speech generation. By allowing a single model to dynamically adjust its codebook size, RAQ can reduce the need for multiple fixed-rate models, thereby potentially lowering the overall computational footprint and energy consumption. These environmental benefits become more pronounced in large-scale or real-time streaming scenarios, where switching between separate models is both costly and carbon-intensive.

As with any generative model, practitioners must remain vigilant about biases in the training data, as such biases could be amplified by RAQ's discrete representations. Nonetheless, we believe the rate flexibility and computational efficiency of RAQ can serve as a net positive contribution, provided that ethical guidelines and transparency measures are carefully observed.


\newpage
\appendix

\section{Appendix / Supplementary Material} \label{appendix}

\subsection{Model-based RAQ: Codebook Clustering} \label{appendix_clustering}
Given a set of the original codebook representations $\mathbf{e}=\{e_i\}_{i=1}^K$, we aim to partition the $K$ codebook vectors into $\widetilde{K}(\leq K)$ codebook vectors $\mathbf{\tilde{e}}=\{\tilde{e}_i\}_{i=1}^{\widetilde{K}}$. Each codebook vector resides in a $D$-dimensional Euclidean space. Using the codebook assignment function $g(\cdot)$, then $g(e_i)=j$ means $i$-th given codebook assigned $j$-th clustered codebook. Our objective for codebook clustering is to minimize the discrepancy $\mathcal{L}$ between the given codebook $\mathbf{e}$ and clustered codebook $\mathbf{\tilde{e}}$:
\begin{equation}
\label{obj_kmean}
\underset{\mathbf{\tilde{e}}, g}{\arg\min} \ \mathcal{L}(\mathbf{e}; \mathbf{\tilde{e}}) =\underset{\mathbf{\tilde{e}}, g}{\arg\min} \sum_{i=1}^{\widetilde{K}} ||e_i-\tilde{e}_{g(e_i)}|| 
\end{equation}
with necessary conditions
\begin{equation}
g(e_i) = \underset{j \in 1, 2, ..., \widetilde{K}}{\arg\min} ||e_i - \tilde{e}_j || \ , \quad \tilde{e}_j=\frac{\sum_{i:g(e_i)=j}e_i}{N_j}
\end{equation}
where $N_j$ is the number of samples assigned to the codebook $\tilde{e}_j$. 

\subsection{Experiment Details} \label{experiment_details}
\subsubsection{Architectures and Hyperparameters}
The model architecture for this study is based on the conventional VQ-VAE framework outlined in the original VQ-VAE paper \citep{van2017neural}, and is implemented with reference to the VQ-VAE-2 \citep{razavi2019generating} implementation repositories \footnote{https://github.com/mattiasxu/VQVAE-2}\footnote{https://github.com/rosinality/vq-vae-2-pytorch}\footnote{https://github.com/EugenHotaj/pytorch-generative}. We are using the ConvResNets from the repositories. These networks consist of convolutional layers, transpose convolutional layers and ResBlocks. Experiments were conducted on two different computer setups: a server with 4 RTX 4090 GPUs and a machine with 2 RTX 3090 GPUs. PyTorch \citep{paszke2019pytorch}, PyTorch Lightning \citep{falcon2019pytorch}, and the AdamW \citep{DBLP:conf/iclr/LoshchilovH19} optimizer were used for model implementation and training. Evaluation metrics such as the Structural Similarity Index (SSIM) and the Frechet Inception Distance (rFID) were computed using implementations of pytorch-msssim \footnote{https://github.com/VainF/pytorch-msssim} and pytorch-fid \footnote{https://github.com/mseitzer/pytorch-fid}, respectively. The detailed model parameters are shown in Table \ref{appendix_params}. RAQs are constructed based on the described VQ-VAE parameters with additional consideration of each parameter.

\begin{table*}[!ht]
\caption{Architecture and hyperparameters  for training VQ-VAE and RAQ models.}
\label{appendix_params}
\vskip 0.15in
\begin{center}
\begin{small}
\scalebox{0.85}{
\begin{tabular}{l c c c c}
\toprule
\textbf{Method} &                \textbf{Parameter}              & \textbf{CIFAR10}      & \textbf{CelebA}   & \textbf{ImageNet}\\
\midrule                        
\multirow{12}{*}{VQ-VAE \citep{van2017neural}}  &                     Input size                    & 32$\times$32$\times$3      & $64\times64\times3$, $128\times128\times3$ & 224$\times$224$\times$3 \\
            &                     Latent layers                 &  8$\times$8      & 16$\times$16, 32$\times$32 & 56$\times$56\\
            &                     Hidden units                  & 128                   & 128  & 256          \\
            &                     Residual units                & 64                    & 64     & 128        \\
            &                     \# of ResBlock                & 2                     & 2     & 2         \\
            &                     Original codebook size ($K$)      & $2^4$ $\sim$ $2^{10}$        & $2^5$ $\sim$ $2^{11}$  & $2^7$ $\sim$ $2^{12}$\\
            &                     Codebook dimension ($d$) & 64                    & 64  & 128           \\
            &                     $\beta$ (Commit loss weight)  & 0.25                  & 0.25     & 0.25  \\
            &                     Weight decay in EMA ($\gamma$) & 0.99                  & 0.99   & 0.99     \\
            &                     Batch size                    & 128                   & 128  & 32     \\
            &                     Optimizer                     & AdamW                 & AdamW  & AdamW    \\
            &                     Learning rate                 & 0.0005                & 0.0005 & 0.0005   \\
            &                     Max. training steps           & 195K                & 635.5K & 961K\\
\midrule
\multirow{5}{*}{\textbf{Model-based RAQ}}  & Original codebook size ($K$)      & 64, 128        & 128, 256 & 512\\
                                &  Adapted codebook size ($\widetilde{K}$)  & $2^{4}$ $\sim$ $2^{10}$ & $2^{5}$ $\sim$ $2^{11}$ & $2^{6}$ $\sim$ $2^{12}$  \\
                                &  Max. DKM iteration                      & 200            & 200 & 200       \\
                                &  Max. IKM iteration                     & 5000           & 5000  & 5000     \\
                                &  $\tau$ of softmax                       & 0.01           & 0.01  & 0.01    \\
\midrule    
\multirow{7}{*}{\textbf{RAQ}} & Original codebook size ($K$)      & 64, 128        & 128, 256 & 512\\
                                &  Adapted codebook size ($\widetilde{K}$)  & $2^{4}$ $\sim$ $2^{10}$ & $2^{5}$ $\sim$ $2^{11}$ & $2^{6}$ $\sim$ $2^{12}$  \\
                                &  Max. Codebook size                      & 1024           & 2048   & 4096           \\
                                &  Min. Codebook size                      & 8              & 16    & 64   \\
                                &  Input size (Seq2Seq)                    & 64             & 64    & 128      \\
                                &  Hidden size (Seq2Seq)                   & 64             & 64    & 128      \\
                                &  \# of reccuruent layers (Seq2Seq)       & 2              & 2      & 2             \\
\bottomrule
\end{tabular}}
\end{small}
\end{center}
\vskip -0.1in
\end{table*}

\subsubsection{Datasets and Preprocessing}
For the \textbf{CIFAR10} dataset, the training set is preprocessed using a combination of random cropping and random horizontal flipping. Specifically, a random crop of size $32\times32$ with padding of 4 using the 'reflect' padding mode is applied, followed by a random horizontal flip. The validation and test sets are processed by converting the images to tensors without further augmentation. For the \textbf{CelebA} dataset, the training set is preprocessed with a series of transformations. The images are resized and center cropped to $64\times64$ or $128 \times 128$, normalized, and subjected to random horizontal flipping. A similar preprocessing is applied to the validation set, while the test set is processed without augmentation. For the \textbf{ImageNet} dataset, the training set is preprocessed with a series of transformations. The images are resized $256\times256$ and center cropped to $224\times224$, normalized, and subjected to random horizontal flipping. A similar preprocessing is applied to the validation set, while the test set is processed without augmentation. These datasets are loaded into PyTorch using the provided data modules, and the corresponding data loaders are configured with the specified batch sizes and learning rate for efficient training (described in Table \ref{appendix_params}. The datasets are used as input for training, validation, and testing of the VQ-VAE model.

\subsubsection{Model Complexity} \label{appendix_model_complexity}

In this section, we provide a comparison of model complexity in terms of the total number of trainable parameters for our VQ-based models, both with and without our RAQ module. The following tables list parameter counts for the VQ-VAE and RAQ (our proposed method) variations on (i) CIFAR10 (Table~\ref{appendix_complexity_cifar10}), (ii) CelebA (Table~\ref{appendix_complexity_celeba}), and (iii) ImageNet (Table~\ref{appendix_complexity_vqgan}).

As shown in the tables below, the addition of RAQ introduces a new Seq2Seq component to facilitate codebook adaptation, resulting in a modest increase in the number of parameters:
\begin{itemize}
    \item In \textbf{VQ-VAE}, whose total parameter count is on the order of a few hundred thousand, the RAQ overhead typically adds \(\sim200\)K+ parameters (e.g., for CIFAR10, the total parameter count increases from about 468K to 732K for \(K=128\)).
    \item In \textbf{stage-1 VQGAN}, which already has tens of millions of parameters, the additional parameters introduced by RAQ are less than 1\% of the total. For instance, the total grows from 72.0M to approximately 72.6M—a practically negligible difference in large-scale settings.
\end{itemize}
These observations demonstrate that RAQ remains practical for a variety of model scales and does not incur substantial overhead, even when applied to deeper architectures.

\begin{table*}[ht]
\caption{Number of parameters for training VQ-VAE and RAQ models on the CIFAR10 dataset.}
\label{appendix_complexity_cifar10}
\begin{center}
\vskip 0.15in
\begin{small}
\scalebox{0.9}{
\begin{tabular}{l ccccc}
\toprule
\textbf{Method}                         & \multicolumn{5}{c}{\# params}                                 \\ 
                                        & Encoder & Decoder    & Quantizer  & Seq2Seq  & Total     \\
\midrule
\textbf{VQ-VAE} ($K=1024$) & 196.3K  & 262K      & 65.5K     & -         & 525K \\
\midrule
\textbf{VQ-VAE} ($K=512$) & 196.3K  & 262K      & 32.8K     & -         & 492K \\
\midrule
\textbf{VQ-VAE} ($K=256$) & 196.3K  & 262K      & 16.4K     & -         & 476K \\
\midrule
\textbf{VQ-VAE} ($K=128$) & 196.3K  & 262K      & 8.2K      & -         & 468K \\
\midrule
\textbf{VQ-VAE} ($K=64$)  & 196.3K  & 262K      & 4.1K      & -         & 463K \\
\midrule
\textbf{VQ-VAE} ($K=32$)  & 196.3K  & 262K      & 2.0K      & -         & 461K \\
\midrule
\textbf{VQ-VAE} ($K=16$)  & 196.3K  & 262K      & 1.0K      & -         & 460K \\
\midrule
\textbf{RAQ} ($K=128$)    & 196.3K  & 262K      & 8.2K      & 263.7K    & 732K \\
\midrule
\textbf{RAQ} ($K=64$)     & 196.3K  & 262K      & 4.1K      & 263.7K    & 728K \\
\bottomrule
\end{tabular}
}
\end{small}
\vskip -0.1in
\end{center}
\end{table*}

\begin{table*}[ht]
\caption{Number of parameters for training VQ-VAE and RAQ models on the CelebA dataset.}
\label{appendix_complexity_celeba}
\begin{center}
\vskip 0.15in
\begin{small}
\scalebox{0.9}{
\begin{tabular}{l ccccc}
\toprule
\textbf{Method}                         & \multicolumn{5}{c}{\textbf{\# params}}                                 \\ 
                                        & Encoder & Decoder    & Quantizer  & Seq2Seq  & Total     \\
\midrule
\textbf{VQ-VAE} ($K=2048$) & 196.3K  & 262K      & 131K     & -         & 590K \\
\midrule
\textbf{VQ-VAE} ($K=1024$) & 196.3K  & 262K      & 65.5K    & -         & 525K \\
\midrule
\textbf{VQ-VAE} ($K=512$)  & 196.3K  & 262K      & 32.8K    & -         & 492K \\
\midrule
\textbf{VQ-VAE} ($K=256$)  & 196.3K  & 262K      & 16.4K    & -         & 476K \\
\midrule
\textbf{VQ-VAE} ($K=128$)  & 196.3K  & 262K      & 8.2K     & -         & 468K \\
\midrule
\textbf{VQ-VAE} ($K=64$)   & 196.3K  & 262K      & 4.1K     & -         & 463K \\
\midrule
\textbf{VQ-VAE} ($K=32$)   & 196.3K  & 262K      & 2.0K     & -         & 461K \\
\midrule
\textbf{RAQ} ($K=256$)     & 196.3K  & 262K      & 16.4K    & 263.7K    & 740K \\
\midrule
\textbf{RAQ} ($K=128$)     & 196.3K  & 262K      & 8.2K     & 263.7K    & 732K \\
\bottomrule
\end{tabular}
}
\end{small}
\vskip -0.1in
\end{center}
\end{table*}

\begin{table*}[!ht]
\caption{Number of parameters for training stage-1 VQGAN and RAQ models on the ImageNet dataset.}
\label{appendix_complexity_vqgan}
\begin{center}
\vskip 0.15in
\begin{small}
\scalebox{0.9}{
\begin{tabular}{l ccc}
\toprule
\textbf{Method}                 & \multicolumn{3}{c}{\textbf{\# params}} \\ 
                                & Quantizer     & Seq2Seq   & Total     \\
\midrule
\textbf{VQGAN} ($K=512$)        & 65.5K         & -         & 72.0M     \\
\midrule
\textbf{VQGAN} ($K=256$)        & 32.8K         & -         & 72.0M     \\
\midrule
\textbf{VQGAN} ($K=128$)        & 16.4K         & -         & 72.0M     \\
\midrule
\textbf{VQGAN} ($K=64$)         & 8.2K          & -         & 72.0M     \\
\midrule
\textbf{VQGAN} ($K=32$)         & 4.1K          & -         & 71.9M     \\
\midrule
\textbf{RAQ} ($K=128$)          & 16.4K         & 610K      & 72.6M     \\
\bottomrule
\end{tabular}
}
\end{small}
\vskip -0.1in
\end{center}
\end{table*}

\subsubsection{Training/Inference Time} 
\label{appendix_time}

We measure training and inference times for our RAQ models on an NVIDIA RTX 3090 GPU with a batch size of 128. We compare \emph{RAQ ($K=256$)} to multiple fixed-rate VQ models ($K\in\{64,256,1024\}$), as well as a model-based RAQ.

Table~\ref{appendix_training_time} and Table~\ref{appendix_inference_time} summarizes the per-epoch training and inference times. As expected, RAQ introduces a Seq2Seq module for codebook adaptation, which leads to a substantial increase in training time compared to a single fixed-rate VQ. However, managing multiple separate VQ models (one for each bitrate) can result in an even higher overall cost in terms of both storage and maintenance, making RAQ more practical in scenarios demanding multiple rates.

At inference, \textbf{our current implementation regenerates the codebook for each mini-batch}, incurring an overhead of approximately $164\%$--$1785\%$ over a fixed-rate VQ. While these overhead numbers appear large, we emphasize that in many real-world use cases (e.g., streaming at a fixed target rate for several frames or batches), one can precompute the adapted codebook once and reuse it for subsequent samples. This caching strategy greatly amortizes the cost of codebook generation, reducing the overhead to a negligible level in practice.

Overall, these observations confirm that RAQ remains computationally feasible while offering flexible, multi-rate compression in a single model. In real-world deployment or streaming setups, the overhead from occasionally generating a codebook is offset by eliminating the need to store or switch between multiple distinct VQ networks.

\begin{table*}[ht]
\caption{Training time per epoch on the CelebA train set using an NVIDIA RTX 3090 GPU.}
\label{appendix_training_time}
\begin{center}
\vskip 0.15in
\begin{small}
\begin{tabular}{l ccc}
\toprule
\textbf{Method}                                     & $K$       & \textbf{Training time per epoch (s)}   & \textbf{\# params}   \\
\midrule
\textbf{VQ-VAE} / \textbf{Model-based RAQ}          & 64        & $18.09 \pm 0.256$                     & 463K   \\
\textbf{VQ-VAE} / \textbf{Model-based RAQ}          & 256       & $18.43 \pm 0.10$                      & 476K   \\
\textbf{VQ-VAE} / \textbf{Model-based RAQ}          & 1024      & $21.64 \pm 0.11$                      & 525K   \\
\textbf{RAQ}                                        & 256       & $514.97 \pm 8.17$                     & 740K   \\
\bottomrule
\end{tabular}
\end{small}
\end{center}
\end{table*}

\begin{table*}[ht]
\caption{Inference time per epoch on the CelebA test set using an NVIDIA RTX 3090 GPU.}
\label{appendix_inference_time}
\begin{center}
\vskip 0.15in
\begin{small}
\begin{tabular}{l cc}
\toprule
\textbf{Method}                             & $\widetilde{K}$     & \textbf{Inference time per epoch (s)}   \\
\midrule
\textbf{VQ-VAE} ($K=64$)                    & -                   & $1.86 \pm 0.10$                        \\
\textbf{VQ-VAE} ($K=256$)                   & -                   & $1.91 \pm 0.12$                         \\
\textbf{VQ-VAE} ($K=1024$)                  & -                   & $1.86 \pm 0.09$                        \\
\textbf{Model-based RAQ} ($K=256$)          & 64                  & $1.98 \pm 0.09$                        \\
\textbf{RAQ} ($K=256$)                      & 64                  & $3.05 \pm 0.11$                        \\
\textbf{Model-based RAQ} ($K=256$)          & 1024                & $70.91 \pm 11.82$                      \\
\textbf{RAQ} ($K=256$)                      & 1024                & $33.21 \pm 0.27$                       \\
\bottomrule
\end{tabular}
\end{small}
\end{center}
\vskip -0.1in
\end{table*}

\subsection{Additional Experiments}

\subsubsection{Effectiveness of \textit{Cross-forcing}} \label{appendix_cross_forcing}

We performed an ablation study to analyze the impact of our \emph{cross-forcing} training strategy on the stability and fidelity of codebook generation. Using a base RAQ model with an original codebook size $K=128$ on the CelebA dataset, we compared two variants: 
(1) \emph{RAQ-w/o-CF} without cross-forcing, and 
(2) \emph{RAQ-w/-CF} with cross-forcing.
Table~\ref{result_cross_forcing} shows the reconstruction metrics (MSE, PSNR, rFID, and SSIM) for different adapted codebook sizes $\widetilde{K}\in\{32,64,128,256,512,1024,2048\}$. We find that \emph{RAQ-w/-CF} gives significantly better performance than \emph{RAQ-w/o-CF} when $\widetilde{K} > K$, leading to up to $4.9\%$ improvement in rFID and noticeable gains in PSNR and SSIM. In contrast, for smaller or equal codebook sizes ($\widetilde{K} \le K$), \emph{RAQ-w-CF} sometimes underperforms its counterpart by a small margin. We hypothesize that cross-forcing is specifically designed to stabilize the generation of \emph{larger} adapted codebooks (up to twice the original size), which can result in a slight tradeoff when quantizing at or below the baseline codebook size.

\begin{table}[!tb]
\caption{Reconstruction performance of RAQ ($K=128$) \textbf{with} or \textbf{without} cross-forcing on the CelebA test dataset}
\label{result_cross_forcing}
\vskip 0.15in
\begin{center}
\begin{small}
\begin{tabular}{lccccc}
\toprule
\textbf{Method}         & $\widetilde{K}$           & MSE $\downarrow$          & PSNR $\uparrow$           & rFID $\downarrow$          & SSIM $\uparrow$ \\
\midrule        
                            & 2048 ($\uparrow$) & \textbf{1.618$\pm$0.016} & \textbf{27.91$\pm$0.04}   & \textbf{22.64$\pm$0.76} & \textbf{0.8810$\pm$0.0013} \\ 
                            & 1024 ($\uparrow$) & \textbf{1.794$\pm$0.027} & \textbf{27.47$\pm$0.07}   & \textbf{24.67$\pm$0.80} & \textbf{0.8710$\pm$0.0016} \\
                            & 512 ($\uparrow$)  & \textbf{2.042$\pm$0.021} & \textbf{26.90$\pm$0.05}   & \textbf{26.90$\pm$0.04} & \textbf{0.8589$\pm$0.0044} \\ 
RAQ-\textbf{w/}-CF    & 256 ($\uparrow$)  & \textbf{2.412$\pm$0.101} & \textbf{26.18$\pm$0.18}   & \textbf{30.81$\pm$1.59} & 0.8391$\pm$0.0125 \\ 
                            & 128 (-)  & 2.801$\pm$0.039           & 25.53$\pm$0.06           & 36.30$\pm$1.12          & 0.8209$\pm$0.0072 \\ 
                            & 64 ($\downarrow$)   & 3.895$\pm$0.095           & 24.10$\pm$0.11           & 47.63$\pm$5.82          & 0.7892$\pm$0.0067 \\ 
                            & 32 ($\downarrow$)   & \textbf{5.357$\pm$0.630}  & \textbf{22.74$\pm$0.54}  & \textbf{62.39$\pm$3.76} & \textbf{0.7414$\pm$0.0304} \\
\midrule        
                            & 2048 ($\uparrow$)     & 1.661$\pm$0.056           & 27.80$\pm$0.14          & 23.58$\pm$0.26          & 0.8789$\pm$0.0030  \\ 
                            & 1024 ($\uparrow$)     & 1.815$\pm$0.050           & 27.42$\pm$0.12          & 25.46$\pm$0.26          & 0.8705$\pm$0.0024  \\ 
                            & 512 ($\uparrow$)      & 2.068$\pm$0.059           & 26.85$\pm$0.12          & 27.81$\pm$0.42          & 0.8567$\pm$0.0046  \\ 
RAQ-\textbf{w/o}-CF  & 256 ($\uparrow$)      & 2.449$\pm$0.052           & 26.12$\pm$0.09          & 32.32$\pm$1.20          & \textbf{0.8407$\pm$0.0031} \\ 
                            & 128 (-)               & \textbf{2.779$\pm$0.015}  & \textbf{25.57$\pm$0.02} & \textbf{36.08$\pm$0.98} & \textbf{0.8261$\pm$0.0019} \\ 
                            & 64 ($\downarrow$)     & \textbf{3.860$\pm$0.237}  & \textbf{24.15$\pm$0.26} & \textbf{45.13$\pm$2.79} & \textbf{0.7942$\pm$0.0154} \\ 
                            & 32 ($\downarrow$)     & 6.289$\pm$0.709           & 22.04$\pm$0.47          & 72.85$\pm$16.69         & 0.7338$\pm$0.0225  \\ \hline

\bottomrule
\end{tabular}
\end{small}
\end{center}
\vskip -0.1in
\end{table}

\subsubsection{Model-based RAQ} \label{appendix_model_based_raq}

\paragraph{Reducing the Rate}

\begin{table}[!t]
\caption{Reconstuction performances of model-based RAQ for \textbf{rate-reduction task} according to adapted codebook size $\widetilde{K}$.}
\label{result_compression}
\vskip 0.15in
\begin{center}
\begin{small}
\begin{tabular}{l|c|ccc}
\toprule
\multirow{2}{*}{\textbf{Method}}        & \multirow{2}{*}{$\widetilde{K}$} &\multicolumn{3}{c}{\textbf{CIFAR10}  ($K=1024$)} \\
                                        &                                  &    PSNR $\uparrow$  & rFID $\downarrow$          & Perplexity $\uparrow$ \\
\midrule
\textbf{VQ-VAE} (baseline model)           & -               & 25.48        &  51.90             & 708.60         \\
\midrule
                     & 512                & 24.35        &  63.67             &\textbf{289.29}         \\ 
\textbf{VQ-VAE} (random select)                          & 256                & 22.81        &  78.00             & 111.77         \\ 
                                        & 128                & 20.87        &  93.57             & 48.87          \\  
\midrule                                        
                                        & 512                &  \textbf{24.62}        &  \textbf{55.78}              & 285.68        \\
\textbf{Model-based RAQ}          & 256                &  \textbf{23.81}        &  \textbf{62.53}              &  \textbf{134.54}         \\
                                        & 128                &  \textbf{23.07}        &  \textbf{69.45}              &  \textbf{73.17}          \\
\midrule
\midrule
\multirow{2}{*}{\textbf{Method}}        & \multirow{2}{*}{$\widetilde{K}$} &\multicolumn{3}{c}{\textbf{CelebA}  ($K=2048$)} \\
                                        &                                  &    PSNR $\uparrow$       & rFID $\downarrow$          & Perplexity $\uparrow$ \\
\midrule
\textbf{VQ-VAE} (baseline model)          & -                             & 28.26          &  22.89                 & 273.47         \\
\midrule
                     & 1024                             & 24.02          &  38.92                 & \textbf{103.50}         \\ 
\textbf{VQ-VAE} (random select)                          & 512                              & 18.99          &  71.64                 & 49.59         \\ 
                                        & 256                              & 23.54          &  115.12                & 27.86          \\  
\midrule                                        
                                        & 1024                             &  \textbf{26.40}          &  \textbf{31.37}                     & 102.36        \\
\textbf{Model-based RAQ}                           & 512                              &  \textbf{25.24}          &  \textbf{39.07}                     &  \textbf{53.45}         \\
                                        & 256                              &  \textbf{24.36}          &  \textbf{45.54}                     &  \textbf{32.86}          \\
\bottomrule
\end{tabular}
\end{small}
\end{center}
\vskip -0.1in
\end{table}

As analyzed in Section~\ref{quant_eval}, RAQ generally outperforms model-based RAQ, but some rate-reduction results on CIFAR10 show that model-based RAQ performs much more stably than in the codebook increasing task. This indicates that simply clustering codebook vectors, without additional neural models like Seq2Seq, can achieve remarkable performance. In Table \ref{result_compression}, the performance via codebook clustering was evaluated with different original/adapted codebook sizes {$K$: 1024 / $\widetilde{K}$: 512, 256, 128} on CIFAR10 and {$K$: 2048 / $\widetilde{K}$: 1024, 512, 256, 128} on CelebA. The conventional VQ-VAE preserved as many codebooks in the original codebook as in the adapted codebook, while randomly codebook-selected VQ-VAE results remained meaningless. Model-based RAQ adopted this baseline VQ-VAE model and performed clustering on the adapted codebook. Model-based RAQ shows a substantial performance difference in terms of reconstructed image distortion and codebook usage compared to randomly codebook-selected VQ-VAE. Even when evaluating absolute performance, it is intuitive that online codebook representation via model-based RAQ provides some performance guarantees.

\paragraph{Increasing the Rate}
In our proposed RAQ scenario, increasing the codebook size beyond the base size is a more demanding and crucial task than reducing it. The crucial step in building RAQ is to achieve higher rates from a fixed model architecture and compression rate, ensuring usability. Therefore, the codebook increasing task was the main challenge. The Seq2Seq decoding algorithm based on cross-forcing is designed with this intention. In Figure~\ref{fig:result_quant}, the codebook generation performance was evaluated with different original/adapted codebook sizes {$K$: 64, 128 / $\widetilde{K}$: 64, 128, 256, 512, 1024} on CIFAR10 and {$K$: 128, 256 / $\widetilde{K}$: 128, 256, 512, 1024, 2048} on CelebA datasets. RAQ outperforms model-based RAQ in the rate-increasing task and partially outperforms conventional VQ-VAE trained on the same codebook size ($K=\widetilde{K}$). This effect is particularly pronounced on CelebA. However, increasing the difference between the original and adapted codebook sizes leads to a degradation of RAQ performance. This effect is more dramatic for model-based RAQ due to its algorithmic limitations, making its performance less stable at high rates. Improving the performance of model-based RAQ, such as modifying the initialization of the codebook vector, remains a limitation.

\subsubsection{Quantitative Results} \label{appendix_quantitative_results}

\paragraph{VQ-VAE}
In Table \ref{appendix_results_cifar10} and \ref{appendix_results_celeba}, we present additional quantitative results for the reconstruction on CIFAR10 and CelebA datasets. The error indicates a 95.45\% confidence interval based on 4 runs with different training seeds. 

\begin{table}[p]
\caption{\textbf{Reconstruction performance} on CIFAR10 dataset. The 95.45\% confidence interval is provided based on 4 runs with different training seeds.}
\label{appendix_results_cifar10}
\vskip 0.15in
\begin{center}
\begin{small}
\scalebox{0.9}{
\begin{tabular}{l c cc c cc}
\toprule
\multirow{2}{*}{\textbf{Method}}    & \ \ Bit Rate                     & \multicolumn{2}{c}{Codebook Usability}  &  Distortion    & \multicolumn{2}{c}{Perceptual Similarity}   \\ \cmidrule(lr){2-7}
                                    & \ \ \ $\widetilde{K}$ & Usage   & Perplexity     & PSNR        & rFID            & SSIM                \\
\midrule
\textbf{VQ-VAE} ($K=\widetilde{K}$) & 1024  & 972.66$\pm$2.97 & 708.60$\pm$7.04 & 25.48$\pm$0.02 & 51.90$\pm$0.51 & 0.8648$\pm$0.0005 \\
\midrule
\textbf{VQ-VAE} ($K=\widetilde{K}$) & 512  & 507.52$\pm$0.51 & 377.08$\pm$5.92 & 24.94$\pm$0.01 & 56.65$\pm$0.91 & 0.8490$\pm$0.0003 \\
\midrule
\textbf{VQ-VAE} ($K=\widetilde{K}$) & 256  & 256$\pm$0 & 204.43$\pm$4.36 & 24.43$\pm$0.02 & 61.40$\pm$0.78 & 0.8310$\pm$0.0006 \\
\midrule
\textbf{VQ-VAE} ($K=\widetilde{K}$) & 128  & 128$\pm$0 & 106.44$\pm$1.54 & 23.85$\pm$0.01 & 66.70$\pm$1.12 & 0.8096$\pm$0.0009 \\
\midrule
\textbf{VQ-VAE} ($K=\widetilde{K}$) & 64  &  64$\pm$0 & 55.64$\pm$0.27  & 23.24$\pm$0.01 & 74.00$\pm$1.64 & 0.7849$\pm$0.0009 \\
\midrule
\textbf{VQ-VAE} ($K=\widetilde{K}$) & 32 & 32$\pm$0 & 29.25$\pm$0.13 & 22.53$\pm$0.02 & 81.68$\pm$1.01 & 0.7545$\pm$0.0009 \\
\midrule
\textbf{VQ-VAE} ($K=\widetilde{K}$) & 16  &  16$\pm$0 & 15.01$\pm$0.21 & 21.76$\pm$0.01 & 89.75$\pm$0.83 & 0.7156$\pm$0.0024 \\
\midrule
                                    & 1024 & 972.66$\pm$2.97 & 708.60$\pm$7.04  & 25.48$\pm$0.02 & 51.90$\pm$0.51 & 0.8648$\pm$0.0005 \\
                                    & 512&  498.38$\pm$1.85 & 289.29$\pm$16.67  & 24.35$\pm$0.11 & 63.67$\pm$2.49 & 0.8305$\pm$0.0056 \\
\textbf{VQ-VAE}                     & 256 & 253.01$\pm$0.66 & 111.77$\pm$21.53 & 22.81$\pm$0.38 & 78.00$\pm$5.07 & 0.7822$\pm$0.0100 \\
($K=1024$)                          & 128& 127.34$\pm$0.33 & 48.87$\pm$11.31  & 20.87$\pm$0.73 & 93.57$\pm$9.87 & 0.7254$\pm$0.0235 \\
(random select)                           & 64 &64$\pm$0 & 24.31$\pm$5.26  & 19.46$\pm$0.98 & 109.90$\pm$14.20 & 0.6720$\pm$0.0309 \\
                                    & 32  & 32$\pm$0 & 13.50$\pm$1.45  & 17.76$\pm$1.12 & 126.57$\pm$15.89 & 0.6102$\pm$0.0350 \\
\midrule
                                    & 1024  & 979.16$\pm$3.72   & 738.48$\pm$7.39       & 25.18$\pm$0.03    & 54.65$\pm$0.99    & 0.8520$\pm$0.0007 \\
                                    & 512   & 507.47$\pm$0.85   & 387.18$\pm$6.87       & 24.82$\pm$0.02    & 57.57$\pm$0.95    & 0.8417$\pm$0.0005 \\
                                    & 256   & 256$\pm$0         & 207.78$\pm$12.13      & 24.34$\pm$0.02    & 61.76$\pm$1.22    & 0.8274$\pm$0.0008 \\
\textbf{RAQ} ($K=128$)              & 128   &  128$\pm$0        & 107.77$\pm$0.58       & 23.91$\pm$0.0    & 65.37$\pm$0.68    & 0.8132$\pm$0.0011 \\
                                    & 64    &  64$\pm$0         & 55.59$\pm$1.81        & 22.87$\pm$0.04    & 77.49$\pm$2.39    & 0.7770$\pm$0.0036 \\
                                    & 32    & 32$\pm$0          & 27.77$\pm$2.03        & 21.85$\pm$0.15    & 89.38$\pm$4.33    & 0.7356$\pm$0.0064 \\
                                    & 16    &  16$\pm$0         & 14.84$\pm$0.74        & 20.82$\pm$0.09   & 98.93$\pm$4.64   & 0.6918$\pm$0.0033 \\

\midrule
& 1024  & 744.36$\pm$18.74 & 395.23$\pm$2.77  & 24.15$\pm$0.03 & 63.88$\pm$1..26 & 0.8213$\pm$0.0014 \\
& 512  &430.06$\pm$11.58 & 256.23$\pm$7.50  & 24.04$\pm$0.03 & 64.74$\pm$0.96 & 0.8177$\pm$0.0012 \\
\textbf{Model-based RAQ} & 256 & 244.61$\pm$3.13 & 185.02$\pm$3.31 & 23.93$\pm$0.01 & 65.65$\pm$1.12 & 0.8139$\pm$0.0010 \\
($K=128$) & 128  & 128$\pm$0 & 106.44$\pm$1.54  & 23.85$\pm$0.01 & 66.70$\pm$1.12 & 0.8096$\pm$0.0009 \\
 & 64  &  64$\pm$0 & 49.55$\pm$1.29  & 22.85$\pm$0.55 & 72.61$\pm$0.77 & 0.7780$\pm$0.0013 \\
& 32  &   32$\pm$0 & 25.65$\pm$0.76  & 21.88$\pm$0.75 & 82.12$\pm$1.74 & 0.7405$\pm$0.0046 \\
& 16  & 16$\pm$0 & 13.79$\pm$0.06  & 20.89$\pm$0.04 & 95.03$\pm$0.34 & 0.6972$\pm$0.0010 \\
\midrule
                                    & 1024  & 972.14$\pm$6.49 & 725.55$\pm$10.90        & 25.04$\pm$0.01    & 55.34$\pm$1.48    & 0.8487$\pm$0.0012 \\
                                    & 512   & 506.38$\pm$1.23 & 382.43$\pm$10.58        & 24.70$\pm$0.02    & 57.91$\pm$1.42    & 0.8387$\pm$0.0011 \\
                                    & 256   & 255.52$\pm$0.48 & 196.17$\pm$9.95         & 24.25$\pm$0.02    & 61.96$\pm$1.00    & 0.8245$\pm$0.0012 \\
\textbf{RAQ} ($K=64$)               & 128   & 128$\pm$0     & 109.65$\pm$3.50           & 23.71$\pm$0.01    & 66.89$\pm$1.07    & 0.8071$\pm$0.0014 \\
                                    & 64    &  64$\pm$0     & 56.31$\pm$0.46            & 23.23$\pm$0.01    & 71.17$\pm$1.17    & 0.7897$\pm$0.0013 \\
                                    & 32    &  32$\pm$0     &29.62$\pm$0.66             & 21.84$\pm$0.09    & 90.04$\pm$1.44    & 0.7350$\pm$0.0038 \\
                                    & 16    &  16$\pm$0     & 15.11$\pm$0.67            & 20.79$\pm$0.18    & 104.86$\pm$5.91   & 0.6918$\pm$0.0084 \\
\midrule
& 1024  &  706.20$\pm$115.18 & 345.50$\pm$107.06 & 23.65$\pm$0.13 & 70.30$\pm$2.02 & 0.8013$\pm$0.0051 \\
& 512  &  428.39$\pm$12.29 & 231.41$\pm$14.64  & 23.55$\pm$0.04 & 71.01$\pm$1.38 & 0.7988$\pm$0.0005 \\
\textbf{Model-based RAQ}& 256 & 233.75$\pm$4.63 & 140.19$\pm$2.82  & 23.39$\pm$0.05 & 71.72$\pm$1.43 & 0.7935$\pm$0.0012 \\
($K=64$)& 128 & 125.07$\pm$1.58 & 101.16$\pm$16.04  & 23.32$\pm$0.05 & 72.68$\pm$1.47 & 0.7901$\pm$0.0008 \\
& 64  & 64$\pm$0 & 55.64$\pm$0.27  & 23.24$\pm$0.01 & 74.00$\pm$1.64 & 0.7849$\pm$0.0009 \\
& 32  & 32$\pm$0 & 26.21$\pm$0.95 & 22.07$\pm$0.13 & 81.61$\pm$2.26 & 0.7569$\pm$0.0014 \\
& 16  & 16$\pm$0 & 13.59$\pm$0.85 & 20.88$\pm$0.23 & 92.84$\pm$3.30 & 0.7004$\pm$0.0063 \\
\bottomrule
\end{tabular}
}
\end{small}
\end{center}
\vskip -0.1in
\end{table}

\begin{table}[p]
\caption{\textbf{Reconstruction performance} on CelebA dataset. The 95.45\% confidence interval is provided based on 4 runs with different training seeds.}
\label{appendix_results_celeba}
\vskip 0.15in
\begin{center}
\begin{small}
\scalebox{0.9}{
\begin{tabular}{l c cc c cc}
\toprule
\multirow{2}{*}{\textbf{Method}}    & \ \ Bit Rate                     & \multicolumn{2}{c}{Codebook Usability}  &  Distortion    & \multicolumn{2}{c}{Perceptual Similarity}   \\ \cmidrule(lr){2-7}
                                    & \ \ \ $\widetilde{K}$ & Usage   & Perplexity     & PSNR        & rFID            & SSIM                \\
\midrule
\textbf{VQ-VAE} ($K=\widetilde{K}$) & 2048  &  779.07$\pm$8.35 & 273.47$\pm$6.86  & 28.26$\pm$0.03 & 22.89$\pm$0.71 & 0.8890$\pm$0.0027 \\
\midrule
\textbf{VQ-VAE} ($K=\widetilde{K}$) & 1024  & 456.86$\pm$3.53 & 160.35$\pm$2.73  & 27.73$\pm$0.05 & 26.67$\pm$1.43 & 0.8763$\pm$0.0029 \\
\midrule
\textbf{VQ-VAE} ($K=\widetilde{K}$) & 512  &  259.59$\pm$3.99 & 95.09$\pm$1.28  & 27.11$\pm$0.01 & 29.77$\pm$0.95 & 0.8636$\pm$0.0022 \\
\midrule
\textbf{VQ-VAE} ($K=\widetilde{K}$) & 256  &  144.44$\pm$2.49 & 57.86$\pm$0.91  & 26.46$\pm$0.03 & 31.53$\pm$1.01 & 0.8481$\pm$0.0009 \\
\midrule
\textbf{VQ-VAE} ($K=\widetilde{K}$) & 128 & 80.26$\pm$0.99 & 34.98$\pm$0.39  & 25.72$\pm$0.04 & 36.25$\pm$0.98 & 0.8279$\pm$0.0027 \\
\midrule
\textbf{VQ-VAE} ($K=\widetilde{K}$) & 64  &  44.94$\pm$1.03 & 20.04$\pm$0.37  & 24.78$\pm$0.03 & 41.22$\pm$0.77 & 0.7986$\pm$0.0037 \\
\midrule
\textbf{VQ-VAE} ($K=\widetilde{K}$) & 32 &  25.48$\pm$0.69 & 12.69$\pm$0.31  & 23.76$\pm$0.06 & 46.56$\pm$1.97 & 0.7660$\pm$0.0032 \\
\midrule
                    & 2048   & 779.07$\pm$8.35 & 273.47$\pm$6.86  & 28.26$\pm$0.03 & 22.89$\pm$0.71 & 0.8890$\pm$0.0027 \\
\textbf{VQ-VAE}     & 1024  &  384.31$\pm$6.76 & 103.50$\pm$3.28  & 24.02$\pm$1.10 & 38.92$\pm$3.27 & 0.7963$\pm$0.0201 \\
($K=2048$)          & 512      &   210.69$\pm$9.23 & 49.59$\pm$4.54  & 18.99$\pm$1.40 & 71.64$\pm$8.27 & 0.7037$\pm$0.0221 \\
(random select)      & 256   &  115.33$\pm$7.73 & 27.86$\pm$3.39  & 16.33$\pm$0.61 & 115.12$\pm$11.93 & 0.6353$\pm$0.0173 \\
\midrule
& 2048  & 885.53$\pm$6.76 & 347.99$\pm$5.17 & 27.96$\pm$0.14 & 23.02$\pm$0.33 & 0.8858$\pm$0.0033 \\
 & 1024  &  490.86$\pm$4.98 & 187.33$\pm$10.37  & 27.51$\pm$0.13 & 25.08$\pm$0.23 & 0.8758$\pm$0.0036 \\
 & 512  &  275.84$\pm$1.72 & 104.61$\pm$5.00  & 26.95$\pm$0.086 & 27.96$\pm$0.49 & 0.8637$\pm$0.0045 \\
\textbf{RAQ} ($K=256$) & 256 &  144.79$\pm$1.21 & 52.63$\pm$0.28  & 26.29$\pm$0.054 & 32.34$\pm$0.86 & 0.8463$\pm$0.0030 \\
& 128  &  80.21$\pm$4.27 & 32.23$\pm$3.87  & 25.13$\pm$0.26 & 39.67$\pm$2.29 & 0.8162$\pm$0.0071 \\
 & 64  & 42.93$\pm$1.61 & 20.85$\pm$1.22 & 24.09$\pm$0.21 & 51.57$\pm$6.66 & 0.7912$\pm$0.0094 \\
 & 32  &  22.76$\pm$1.57 & 12.32$\pm$0.91  & 22.62$\pm$0.27 & 69.65$\pm$9.49 & 0.7479$\pm$0.0129 \\
\midrule
 & 2048  & 704.17$\pm$108.04 & 117.53$\pm$33.57  & 26.54$\pm$0.10 & 30.34$\pm$1.39 & 0.8507$\pm$0.0041 \\
 & 1024  &460.77$\pm$26.98 & 134.48$\pm$11.26  & 26.59$\pm$0.06 & 30.49$\pm$1.10 & 0.8509$\pm$0.0021 \\
& 512  &  279.53$\pm$9.48 & 100.64$\pm$08.94  & 26.40$\pm$0.08 & 30.95$\pm$0.98 & 0.8488$\pm$0.0017 \\
 \textbf{Model-based RAQ} & 256 &   144.44$\pm$2.49 & 57.86$\pm$0.91  & 26.46$\pm$0.03 & 31.53$\pm$1.01 & 0.8481$\pm$0.0009 \\
 ($K=256$) & 128 &  75.31$\pm$3.09 & 25.05$\pm$1.95  & 24.44$\pm$0.25 & 38.95$\pm$2.91 & 0.7890$\pm$0.0141 \\
 & 64  &  41.66$\pm$1.22 & 14.73$\pm$0.56 3 & 22.85$\pm$0.36 & 48.96$\pm$1.13 & 0.7391$\pm$0.0192 \\
 & 32 &  22.96$\pm$0.90 & 10.16$\pm$0.95 & 21.81$\pm$0.45 & 62.46$\pm$0.00 & 0.7077$\pm$0.0195 \\
\midrule
& 2048  & 891.13$\pm$7.11 & 345.25$\pm$5.15 & 27.91$\pm$0.04 & 22.64$\pm$0.76 & 0.8810$\pm$0.0013 \\
 & 1024  &  490.15$\pm$14.39 & 176.71$\pm$6.19  & 27.47$\pm$0.07 & 24.67$\pm$0.80 & 0.8710$\pm$0.0016 \\
& 512 &  272.60$\pm$2.08 & 96.87$\pm$2.68 & 26.90$\pm$0.05 & 26.90$\pm$0.04 & 0.8589$\pm$0.0044 \\
\textbf{RAQ} ($K=128$)& 256  &  152.65$\pm$2.45 & 60.90$\pm$2.18 & 26.18$\pm$0.18 & 30.81$\pm$1.59 & 0.8391$\pm$0.0125 \\
& 128 &  79.17$\pm$0.93 & 31.36$\pm$0.77  & 25.53$\pm$0.06 & 36.30$\pm$1.12 & 0.8209$\pm$0.0072 \\
 & 64 & 42.71$\pm$1.66 & 19.78$\pm$2.31  & 24.10$\pm$0.11 & 47.63$\pm$5.82 & 0.7892$\pm$0.0067 \\
& 32 &  22.42$\pm$1.92 & 11.43$\pm$2.14 & 22.74$\pm$0.54 & 62.39$\pm$3.76 & 0.7414$\pm$0.0304 \\
\midrule
& 2048  & 350.02$\pm$100.57 & 64.87$\pm$21.22 & 22.77$\pm$0.78 & 52.37$\pm$10.94 & 0.7463$\pm$0.0347 \\
& 1024  &   432.15$\pm$45.80 & 102.79$\pm$17.34  & 25.57$\pm$0.19 & 35.62$\pm$1.46 & 0.8296$\pm$0.0062 \\
 & 512  &   262.78$\pm$29.47 & 75.63$\pm$12.04  & 25.50$\pm$0.29 & 36.82$\pm$0.73 & 0.8265$\pm$0.0026 \\
 \textbf{Model-based RAQ}& 256 & 153.16$\pm$5.46 & 53.22$\pm$4.62  & 25.42$\pm$0.28 & 36.78$\pm$1.27 & 0.8285$\pm$0.0022 \\
  ($K=128$)& 128  &  80.26$\pm$0.99 & 34.98$\pm$0.39  & 25.72$\pm$0.04 & 36.25$\pm$0.98 & 0.8279$\pm$0.0027 \\
 & 64 &  41.88$\pm$0.72 & 16.70$\pm$0.43  & 23.63$\pm$0.16 & 47.09$\pm$4.09 & 0.7736$\pm$0.0080 \\
 & 32  & 23.31$\pm$0.89 & 9.56$\pm$0.77 & 21.64$\pm$0.13 & 64.85$\pm$6.92 & 0.7037$\pm$0.0102 \\
\bottomrule
\end{tabular}
}
\end{small}
\end{center}
\vskip -0.1in
\end{table}

\paragraph{VQ-VAE-2}

\begin{figure}[!t]
\centering
\includegraphics[width=0.9\linewidth]{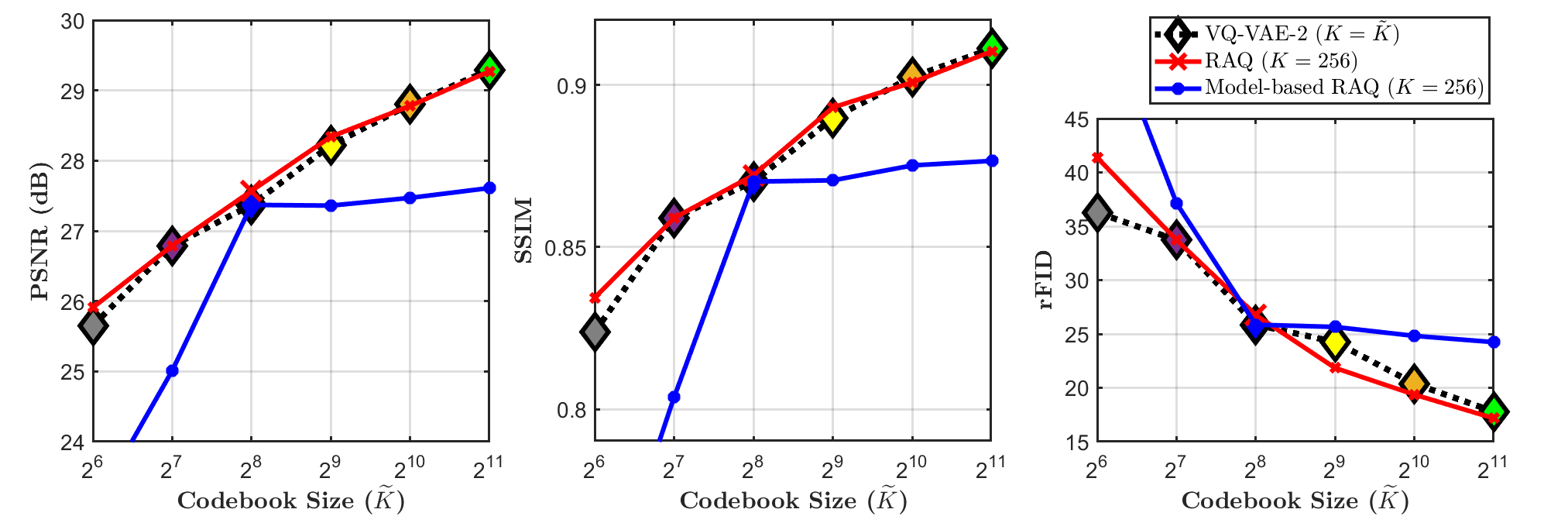}
\caption{\textbf{Reconstruction performance} at different rates (adapted codebook sizes) evaluated on CelebA ($64\times64$) test set. In the graph, the black VQ-VAE-2s \citep{razavi2019generating} are separate models trained on each codebook size, while the RAQs are one model per line.}
\label{fig:result_vq2_appendix}
\end{figure}

Figure~\ref{fig:result_vq2_appendix} shows the reconstruction performance using VQ-VAE-2 as the baseline model.

\subsubsection{Additional Qualitative Results} \label{appendix_qualitative_results}

\begin{figure}[ht]
\begin{center}
\centerline{\includegraphics[width=0.95\linewidth]{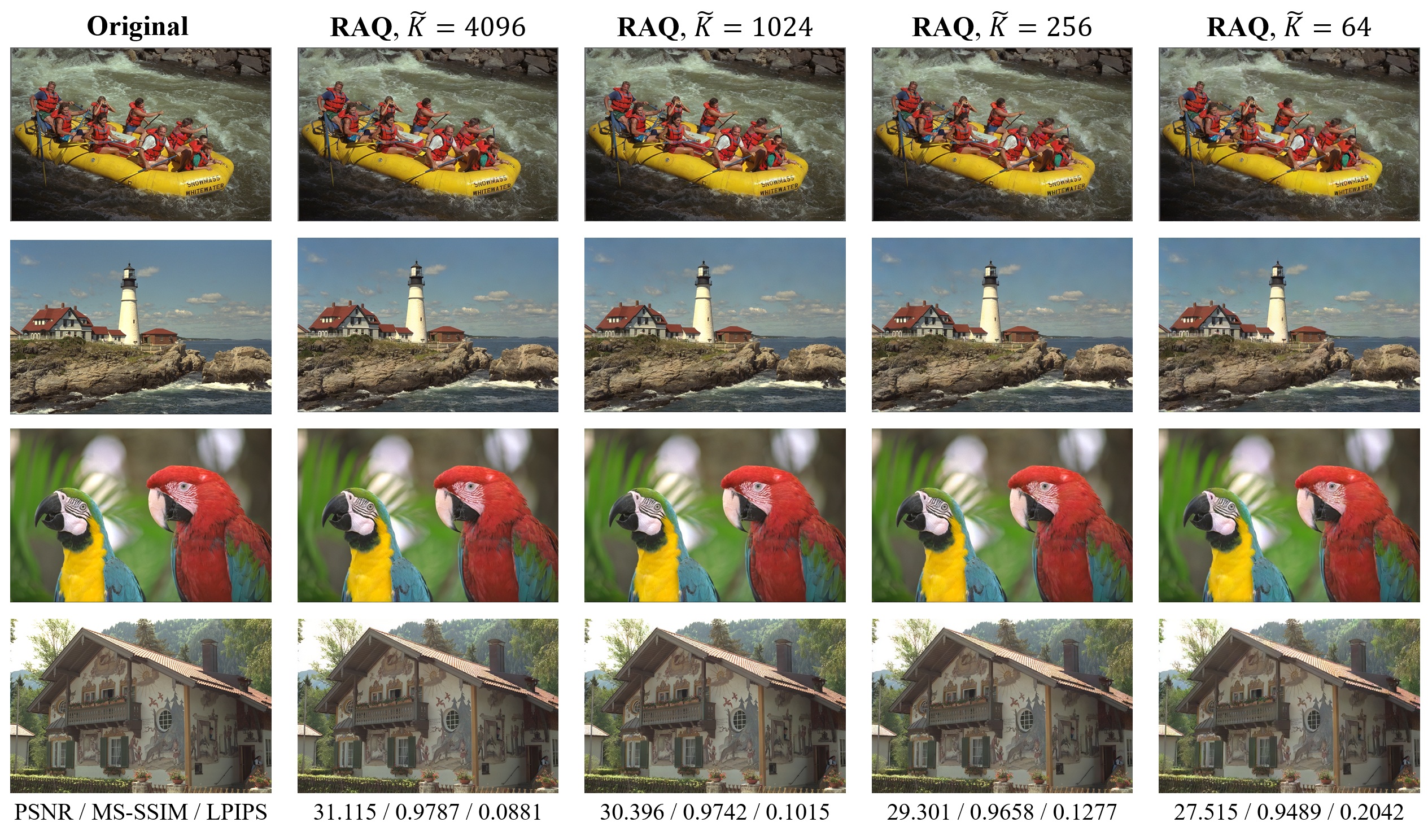}}
\caption{\textbf{Reconstructed images} for Kodak \cite{kodak} dataset at different rates.}
\label{fig:result_quali_kodak}
\end{center}
\vskip -0.2in
\end{figure}

We conducted additional experiments using the VQ-VAE-2 model \citep{razavi2019generating} with an original codebook size of $K=512$. To enhance perceptual quality, we incorporated the LPIPS loss \citep{zhang2018unreasonable} into the training objective and trained the model on the ImageNet dataset at a resolution of $256 \times 256$. The reconstruction task involved reconstructing 24 high-quality images from the Kodak dataset \citep{kodak}, each with a resolution of $768 \times 512$. For codebook adaptation, we adjusted the codebook size to $\widetilde{K} \in \{4096, 1024, 256\}$ using our RAQ framework. The qualitative results are illustrated in Figure~\ref{fig:result_quali_kodak}. Contrary to Figure~\ref{fig:result_quali}, where reducing the codebook size in a less complex VQ-VAE model led to noticeable performance degradation, our RAQ-based VQ-VAE-2 demonstrated robust performance across various codebook sizes. Specifically, even as the codebook size decreased, the RAQ-based VQ-VAE-2 model effectively preserved image quality at higher resolutions. These results indicate that increasing the model complexity and refining the training methodology significantly enhance the RAQ framework's ability to adapt codebook rates without compromising reconstruction fidelity. This demonstrates the superior effectiveness of our method in handling variable-rate compression tasks, particularly in high-resolution image reconstruction scenarios.

\end{document}